\PassOptionsToPackage{table}{xcolor}
\documentclass[11pt]{article}
\usepackage[preprint]{acl}
\usepackage{times}
\usepackage{latexsym}

\usepackage{amsmath,amsfonts,bm}

\def\eqref#1{equation~\ref{#1}}

\def\1{\bm{1}}

\DeclareMathAlphabet{\mathsfit}{\encodingdefault}{\sfdefault}{m}{sl}
\SetMathAlphabet{\mathsfit}{bold}{\encodingdefault}{\sfdefault}{bx}{n}

\usepackage[utf8]{inputenc} 
\usepackage[T1]{fontenc}    
\usepackage{booktabs}       
\usepackage{amsfonts}       
\usepackage{pifont}         
\usepackage{nicefrac}       
\usepackage{microtype}      
\usepackage{amsmath}
\usepackage{cleveref}
\usepackage{float}
\usepackage{makecell}
\usepackage{adjustbox}
\usepackage[linesnumbered,ruled,vlined]{algorithm2e}
\usepackage[most]{tcolorbox}
\usepackage{amssymb}
\usepackage{listings}
\usepackage{graphicx}  
\usepackage{courier}  
\tcbuselibrary{skins, breakable}

\tcbuselibrary{listings, skins} 

\newtcblisting{promptbox}[1]{
    colback=blue!5,           
    colframe=blue!75,         
    fonttitle=\bfseries,      
    title={#1},               
    arc=2pt, outer arc=2pt,   
    boxsep=5pt,               
    left=5pt, right=5pt, top=5pt, bottom=5pt, 
    listing only,             
    listing options={
        language=Python,
        basicstyle=\scriptsize\ttfamily\color{blue}, 
        keywordstyle=\bfseries\color{blue},          
        stringstyle=\color{blue},                    
        commentstyle=\color{blue},                   
        identifierstyle=\color{blue},                
        breaklines=true,                              
        showstringspaces=false,
        mathescape=true,                              
        frame=none,                                   
        xleftmargin=0pt,                              
        aboveskip=0pt, belowskip=0pt,                 
        backgroundcolor={}                            
    }
}

\newtcblisting{plainpromptbox}[1]{
    breakable, enhanced,
    colback=gray!6, colframe=gray!60!black, boxrule=0.6pt,
    colbacktitle=gray!22, coltitle=black,
    fonttitle=\bfseries\footnotesize, title={#1},
    arc=3pt, outer arc=3pt, boxsep=2pt,
    left=5pt, right=5pt, top=3pt, bottom=3pt,
    before skip=6pt, after skip=6pt,
    listing only,
    listing options={
        basicstyle=\scriptsize\rmfamily, breaklines=true, breakindent=0pt,
        breakatwhitespace=false, columns=fullflexible, keepspaces=true,
        showstringspaces=false, frame=none, backgroundcolor={},
        xleftmargin=0pt, aboveskip=0pt, belowskip=0pt
    }
}

\usepackage{tabularx}
\usepackage{subcaption}
\usepackage{dsfont}
\usepackage{multirow}

\usepackage{newunicodechar}
\usepackage{wrapfig}
\usepackage{placeins}
\usepackage{afterpage} 
\usepackage{titletoc}

\definecolor{orange}{HTML}{ffaa76}
\definecolor{pink}{HTML}{e774b6}
\definecolor{blue}{HTML}{3C69CC}
\definecolor{brown}{HTML}{94673E}
\definecolor{yellow}{HTML}{FFCB47}

\newcommand{\rewardgain}[2]{%
  \makebox[\linewidth][c]{#1}%
  \llap{\raisebox{-1.25ex}{\textcolor{green!55!black}{\scalebox{0.72}{(#2)}}}}%
  \rule[-2.0ex]{0pt}{0pt}%
}

\title{Just-in-Time Skill Retrieval via Long-Horizon State Tracking}
\author{
  \textbf{Xin Lin}$^{1}$\thanks{Equal contribution.},
  \textbf{Ziheng Wang}$^{1}$\footnotemark[1],
  \textbf{Zerui Chen}$^{1}$,
  \textbf{Jiayao Ma}$^{1}$,
  \textbf{Zhimin Wei}$^{1}$,
  \\
  \textbf{Bingshuai Liu}$^{1}$,
  \textbf{Buchao Zhan}$^{1}$,
  \textbf{Qinggang Zhang}$^{2}$\thanks{Corresponding authors.},
  \textbf{Jinsong Su}$^{1}$\footnotemark[2]
  \\[2pt]
  $^{1}$Xiamen University
  \qquad
  $^{2}$Jilin University
  \\[2pt]
  \texttt{linxin1@stu.xmu.edu.cn}
  \\
  \texttt{qinggangzhang@jlu.edu.cn}
  \qquad
  \texttt{jssu@xmu.edu.cn}
}

\begin{document}

\maketitle

\begin{abstract}

Large language model (LLM) agents increasingly rely on external skill libraries to support complex task execution. However, determining which skills are needed remains challenging in long-horizon tasks. Existing methods typically select a fixed skill set before execution, implicitly assuming that a task's procedural requirements can be determined from its initial description. This assumption is restrictive for long-horizon interaction, where new requirements emerge from intermediate observations, execution outcomes, and newly revealed constraints. This motivates just-in-time skill retrieval, which dynamically selects skills as execution unfolds. Just-in-time skill retrieval is promising but challenging: the utility of a skill depends not only on the current context, but also on the evolving task state shaped by prior steps. To this end, we propose JitSkill, a state-grounded framework that addresses this challenge through continuous task-state tracking. By connecting emerging requirements with accumulated task progress, JitSkill provides procedural support throughout execution without treating each retrieval decision independently. Experiments across three agent benchmarks demonstrate its effectiveness in improving long-horizon task completion. Our code is available at \textcolor{blue}{\url{https://github.com/XMUDeepLIT/JitSkill.git/}}.

\end{abstract}

\section{Introduction}

\label{sec:introduction}
Large Language Model (LLM)-based agents increasingly rely on external knowledge to support complex reasoning, tool use, and interaction with their environments~\citep{feng2026graph,zhou2026eviskill,huang2026gta,wu2026cosplay,yuan2026osworld2,jang2026odysseys}. 
Among different forms of external knowledge, skills provide a natural representation of procedural experience: they capture how an agent accomplishes recurring goals or operations by encoding successful task executions as reusable procedures that agents can invoke again when similar demands arise~\citep{xia2026skillrl,wu2026cosplay,yu2026masa}.
As these skills accumulate across tasks, external skill libraries provide agents with a growing repository of procedural knowledge that can complement their underlying reasoning and action capabilities~\citep{wang2023voyager,li2026skillsbench,zhang2026skillpoison}.

\begin{figure*}[!t]
    \centering
    \includegraphics[width=\textwidth]{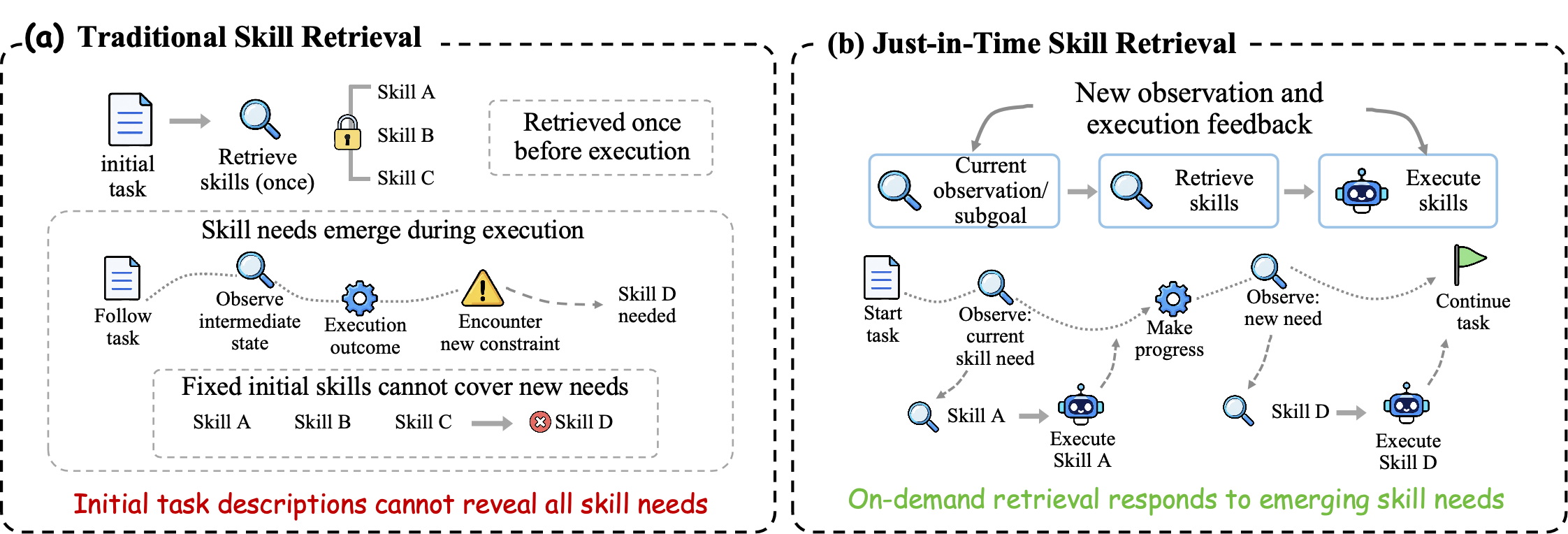}
    \caption{\textbf{Comparison of pre-execution and just-in-time skill retrieval.} \textbf{(a)} Pre-execution skill retrieval can miss needs revealed during execution. \textbf{(b)} Just-in-time retrieval acquires skills on demand as needs become observable.}
\label{fig:intro_skill_retrieval}
\end{figure*}

As skill libraries continue to grow, exposing all available skills to an agent becomes increasingly impractical: it increases the context burden and introduces irrelevant or only partially relevant procedures that may interfere with subsequent reasoning~\citep{su2026sra,li2026skillflow,kang2026skillret}. Existing approaches therefore largely formulate skill selection as a retrieval problem, aiming to identify a compact subset of useful skills from an expanding skill library. Early retrieval-based approaches primarily identify individual skills according to their relevance to the task, as exemplified by SkillFlow~\citep{li2026skillflow} and SkillRouter~\citep{zheng2026skillrouter}. More recent methods move beyond independent skill matching by explicitly modeling relationships among skills: GoS~\citep{liu2026graphofskills}, GoSkills~\citep{zeng2026goskills}, and CaSKG~\citep{li2026caskg} exploit dependency, grouping, or causal relations to retrieve and organize structurally related skills. SkillAligner~\citep{li2026skillaligner} further considers compatibility after retrieval, identifying mismatches among retrieved skills, task requirements, and execution interfaces and adapting skill fragments accordingly.
Together, these advances have progressively extended skill retrieval from individual relevance matching to structured retrieval and alignment, enabling agents to access increasingly large and heterogeneous skill libraries.

Despite these advances, existing retrieval methods largely assume that the skills required for a task can be sufficiently determined from the information available before execution begins. 
This assumption becomes increasingly restrictive in long-horizon interactive tasks, where capability demands are not fully specified by the initial instruction but emerge progressively through intermediate observations, execution outcomes, and newly discovered constraints~\citep{yao2023react,zhou2024webarena}. 
A skill set selected at the outset may therefore provide adequate support for the initial task description while failing to cover procedural needs that arise later in the trajectory.
This limitation has motivated a shift toward \emph{just-in-time skill retrieval}, where skills are retrieved on demand as task execution unfolds. Recent web and computer-use agents instantiate this paradigm by dynamically identifying skill requirements from the current observation, subgoal, or local execution context and retrieving corresponding skills when those requirements emerge~\citep{li2026sgdr,chen2026cuaskill}. Rather than predicting all necessary skills from the initial task description, just-in-time retrieval allows the retrieved skill set to evolve with execution, retrieving skills when their corresponding needs actually become observable.

Just-in-time retrieval is promising but challenging. (i) Whether a skill is needed depends not only on the current context but also on prior execution. The same observation or subgoal can call for different procedural support depending on what has already been accomplished, which resources have been acquired, which attempts have failed, and which requirements remain unresolved~\citep{bai2026skilldag,zhou2026memento,xu2026hyperskill}. A skill that is useful early in a trajectory may become redundant once the requirement it addresses has been satisfied, or even conflict with decisions established later in execution. Consequently, retrieving skills only from the current observation or local subgoal can produce individually reasonable decisions that are inconsistent with the task's accumulated progress. (ii) Maintaining a task state that reliably guides skill retrieval and execution is challenging. 
Simply retaining the full history can obscure the task-relevant consequences of prior interactions, causing the agent to repeat completed work, revisit invalidated paths, or overlook commitments that should constrain subsequent skill decisions.
Together, these challenges create a fundamental risk in just-in-time retrieval: locally relevant skills can be repeatedly selected and successfully executed without forming a coherent trajectory toward the remaining task goals. We refer to this failure as \emph{execution drift}, which manifests as redundant exploration or execution-consistency deviation.

\begin{figure}[!t]
\centering
\begin{minipage}{\linewidth}
\centering
\includegraphics[width=0.97\linewidth]{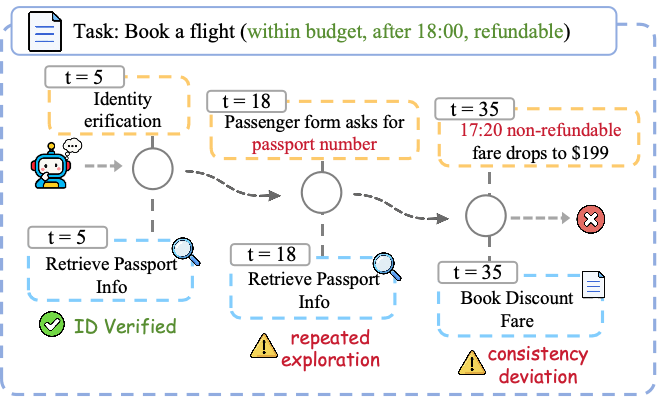}\par
{\small (a)}
\end{minipage}
\par\vspace{3pt}
\begin{minipage}[b]{0.49\linewidth}
\centering
\includegraphics[width=\linewidth]{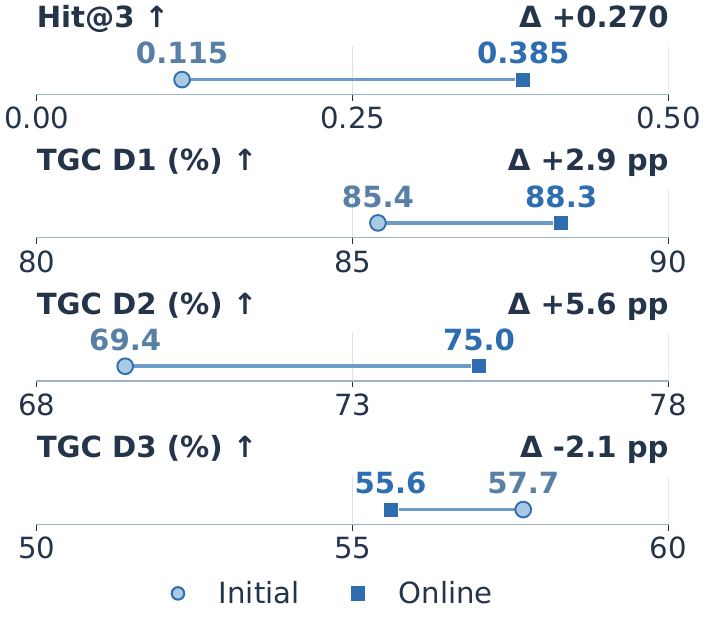}\par
{\small (b)}
\end{minipage}\hfill
\begin{minipage}[b]{0.49\linewidth}
\centering
\includegraphics[width=\linewidth]{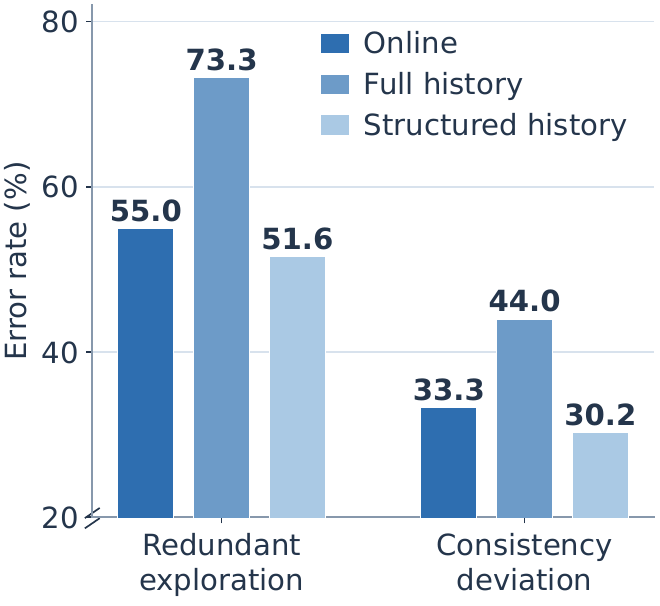}\par
{\small (c)}
\end{minipage}
\caption{\textbf{Preliminary analysis of skill retrieval.} \textbf{(a)} A travel-task example illustrating errors caused by overlooking prior execution progress. \textbf{(b)} Comparison of initial and online retrieval on AppWorld in terms of stage-specific skill coverage (Hit@3) and task goal completion (TGC) by difficulty. \textbf{(c)} Redundant exploration and execution-consistency deviation under local-context, full-history, and structured-history retrieval.}
\label{fig:preliminary-execution-drift}
\end{figure}

To address these challenges, we propose \textsc{JitSkill}, a state-grounded skill retrieval framework that couples just-in-time skill retrieval with long-horizon task-state tracking. Rather than treating each retrieval decision as an independent response to the current context, \textsc{JitSkill} continuously maintains a compact representation of the task state and uses it to guide both skill retrieval and subsequent execution. Specifically, it maintains complementary explicit and latent  representations that together constitute the task state.
The explicit record organizes prior operations, their outcomes, and failure feedback into accessible textual evidence, while the latent representation recurrently summarizes historical interactions that cannot be fully retained in bounded explicit fields.
At each decision point, the skill navigator grounds just-in-time skill retrieval in the tracked task state while responding to the current execution context. The executor receives the selected skills together with the explicit record and latent state tokens to execute the next action. The resulting action and environment feedback subsequently update the task state, allowing their consequences to inform later retrieval decisions.
In this way, task state serves as a persistent bridge between past execution and future skill decisions, allowing skill retrieval and execution to evolve with the trajectory without losing sight of what has already been established and what remains to be accomplished.

Our contributions are as follows:
\begin{itemize}
    \item We identify two limitations of existing skill retrieval paradigms: fixed pre-execution skill retrieval cannot cover demands that emerge as a task progresses, while local online matching cannot determine whether selected skills remain consistent with accumulated task progress. We show why treating each retrieval decision as an independent response to the current context leaves both unresolved.
    \item We introduce \textsc{JitSkill}, a task-state-grounded framework for just-in-time skill retrieval. It tracks task progress through complementary explicit and latent states, which guide on-demand retrieval and are provided to the executor for state-aware skill execution.

    \item Across SkillsBench, ALFWorld, and $\tau^2$-Bench, \textsc{JitSkill} achieves the highest overall task success rates, with a $20.6\%$ relative improvement over the strongest baseline on SkillsBench. Ablation results further support the benefit of jointly grounding skill retrieval and execution in the tracked task progress.
   
\end{itemize}

\section{Preliminary Study}
\label{sec:preliminary}

We examine the limitations of fixed pre-execution skill retrieval in long-horizon tasks, analyze how online retrieval based on local execution context affects task progress, and explore ways to capture task-state information from interaction trajectories by comparing full histories with structured history records.
Experimental details are in Appendix~\ref{app:preliminary-details}.

\paragraph{Execution Requires Continually Updated Skill.}
We compare Initial and Online Retrieval at the same execution decision points using Hit@3, which measures whether the top three retrieved skills include a skill compatible with the stage-specific need. Online Retrieval raises Hit@3 from $0.115$ to $0.385$. This gap indicates that skills retrieved from the initial instruction cannot adequately cover procedural needs that become apparent during execution. Intermediate observations and execution outcomes provide additional information about these needs, motivating retrieval that can be updated as the task unfolds.

\paragraph{Skill Retrieval Based Solely on Local Context Can Cause Execution Drift.}
Figure~\ref{fig:preliminary-execution-drift}a illustrates how similar local cues can require different skill decisions depending on prior progress. A passport-lookup skill helps satisfy a prerequisite before the ID is obtained, but becomes redundant once a valid ID has been saved. Likewise, a cheapest-fare skill may match a current offer while conflicting with established departure-time or refund requirements. Local relevance alone cannot establish whether a skill is needed or consistent with conditions accumulated during execution. At the task level, Online Retrieval improves TGC over Initial Retrieval by $2.9$ and $5.6$ percentage points on D1 and D2, but decreases it by $2.1$ points on D3. Together with the coverage results, this shows that making stage-relevant skills available does not consistently translate into higher task completion, motivating consideration of prior results and unresolved requirements in subsequent skill decisions.

\paragraph{Interaction Histories Require Careful Organization into Task States.}
We compare local-context retrieval with two ways of incorporating prior interactions: the full history and a short structured record. Simply appending the full history does not improve skill retrieval. Instead, it lowers Hit@3 from $0.385$ to $0.346$ while substantially increasing query length, and leads to more execution-consistency deviations (Figure~\ref{fig:preliminary-execution-drift}c). In contrast, a structured record that summarizes the task goal, previously called interfaces, and observed failures reduces both redundant exploration and execution-consistency deviation compared with local-context and full-history retrieval. These findings suggest that the value of interaction history depends not on how much is retained, but on how task-relevant information is organized. This motivates an evolving task state that preserves prior execution outcomes and constraints to guide subsequent skill decisions.

\paragraph{Takeaway.}
These diagnostics motivate three design choices in \textsc{JitSkill}. Skill retrieval should be updated as new needs emerge, interaction history should be organized into an evolving task state, and this state should guide skill acquisition and execution. The state should identify missing procedures, reusable results, failure conditions that warrant avoiding repeated attempts, and  constraints that subsequent local cues should respect. \textsc{JitSkill} realizes this guidance through complementary explicit and latent representations, and provides both of them to a resource-constrained executor.

\section{The Framework of \textsc{JitSkill}}
\label{sec:method}
In this section, we present \textsc{JitSkill}, a framework that grounds just-in-time skill retrieval and subsequent skill execution in a continuously tracked task state. It maintains complementary explicit and latent representations to retain task progress across interactions. We first introduce the framework in Section~\ref{sec:method-overview}, then describe task-state tracking in Section~\ref{sec:state-organization}, and finally explain state-grounded skill retrieval and state-aware execution in Section~\ref{sec:state-retrieval}.

\begin{figure*}[t]
    \centering
    \includegraphics[width=\textwidth]{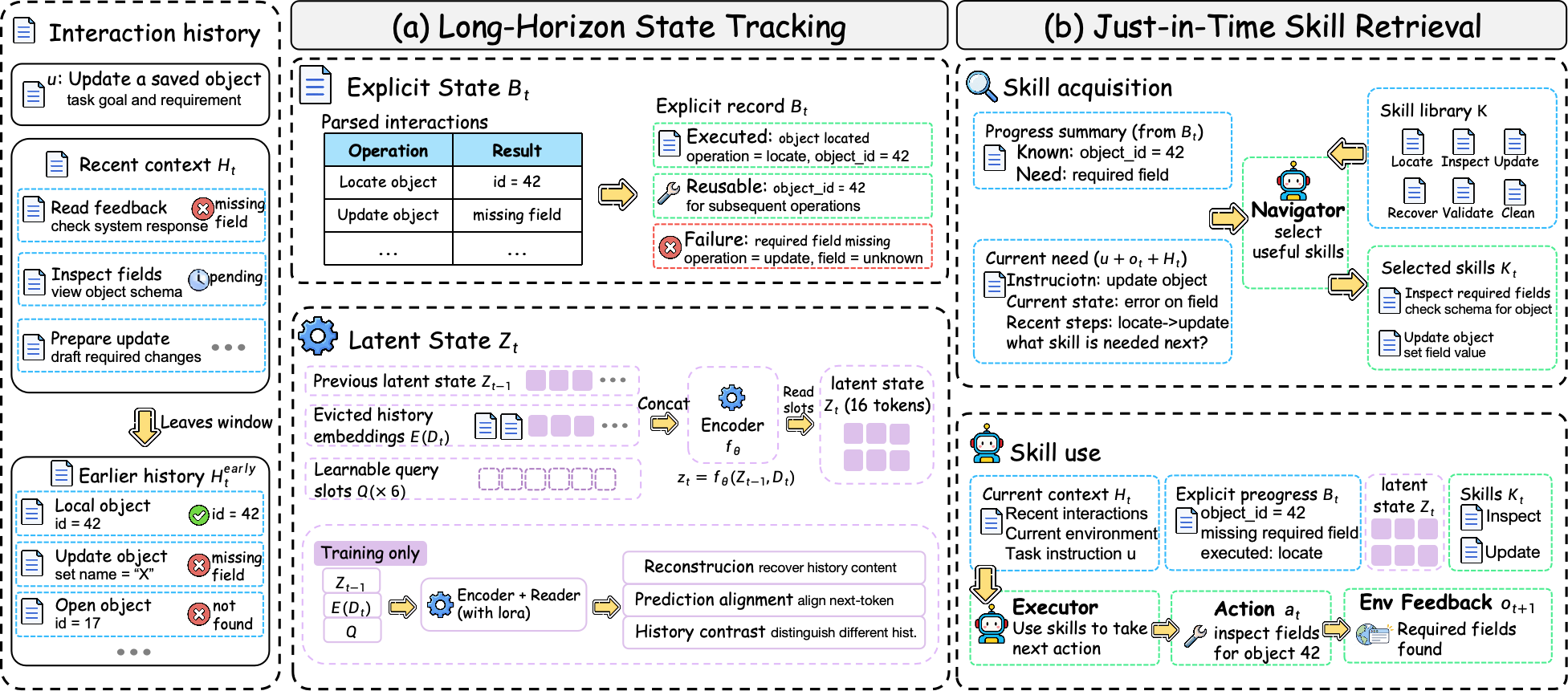}
    \vspace{-5mm}
    \caption{Overview of \textsc{JitSkill}. Explicit records and latent representations jointly track task progress, guiding both skill retrieval and action generation. The executor receives the retrieved skills together with the two state representations. Actions and feedback enter the recent context; accumulated history is organized into explicit records and compressed into latent representations for subsequent decisions.}
    \label{fig:main_pipeline}
    \vspace{-5mm}
\end{figure*}

\subsection{Overview}
\label{sec:method-overview}

We consider long-horizon interactive tasks in which an agent retrieves reusable skills from an external library $\mathcal K=\{k_1,\ldots,k_N\}$. At decision point $t$, the current execution context consists of the task instruction $u$, the observation $o_t$, and a bounded recent interaction history $H_t$. To retain task-relevant execution information beyond this recent context, \textsc{JitSkill} maintains an evolving task state $S_t=(B_t,Z_t)$. The explicit state $B_t$ records inspectable task facts and execution outcomes, while the fixed-size latent state $Z_t$ retains historical context through recurrent updates. Together, they represent the accumulated task progress relevant to subsequent skill decisions across execution steps.

As illustrated in Figure~\ref{fig:main_pipeline}, a skill navigator selects skills from the external library according to the tracked task state and current execution context, producing a skill set $\mathcal K_t\subseteq\mathcal K$ within the skill budget. The executor combines these skills with the current context, the explicit state $B_t$, and the projected latent state tokens $Z'_t$ to generate the next action $a_t$. The action and resulting environment feedback $o_{t+1}$ inform subsequent updates to the task state and recent context, closing the loop between task-state tracking, just-in-time skill retrieval, and state-aware execution over successive task steps.

\subsection{Task-State Tracking}
\label{sec:state-organization}

Task-state tracking organizes execution evidence into complementary representations that provide sufficient context for later decisions. The explicit record organizes discrete task facts into a form that can be directly inspected and updated during execution. The latent state complements these records by continuously carrying forward context distributed across the execution process, which is difficult to predefine as a small set of fixed textual fields.

\paragraph{Explicit task state.}

The explicit state $B_t$ is a bounded structured textual summary that continuously records and updates important information during long-horizon interaction. It preserves inspectable task facts, including key facts required for task completion, important subtasks and their progress, completed operations and their outcomes, and failed attempts with their associated feedback.

To update $B_t$, we extract completed operations, their outcomes, and observed failures from the interaction history preceding the current decision, denoted by $H_{<t}$:
\begin{equation}
B_t=\mathcal E(u,H_{<t}),
\label{eq:explicit-progress}
\end{equation}
where $\mathcal E$ denotes the evidence-extraction procedure.

\paragraph{Latent task state.}

While $B_t$ records discrete and inspectable task facts, $Z_t$ recurrently summarizes historical interactions to retain execution context across task steps~\citep{zhu2026skillmemcompactexternalskill}. \textsc{JitSkill} maintains a fixed number of latent slots for this purpose. After each interaction, we denote the newly accumulated action--feedback block by $D_{t+1}$. The encoder integrates this block with the preceding latent state $Z_t$ to produce the updated state $Z_{t+1}$:
\begin{equation}
Z_{t+1}=f_\theta(Z_t,D_{t+1}),\qquad
Z_t\in\mathbb R^{m\times d},
\label{eq:latent-state}
\end{equation}
\noindent where $m$ is the fixed number of latent slots and $d$ is their dimension. By continually compressing new action--feedback evidence into the same slots, the encoder carries task state forward across interactions in a fixed-size form. This persistent representation supports long-horizon task-state tracking by retaining evolving execution context without repeatedly encoding the full history at each step.

\paragraph{Learning latent representations.}
We freeze the base language model and adapt the encoder with LoRA. We first pretrain the encoder with a history-compression objective $\mathcal L_{\mathrm{comp}}$ that preserves task-relevant information across historical interactions:
\begin{equation}
\begin{aligned}
\mathcal L_{\mathrm{comp}}=\mathbb E_{i,t}\big[&
\omega_{i,t}\mathcal L_{\mathrm{rec}}
+\beta_i\mathcal L_{\mathrm{align}}\\
&+\gamma_i\mathcal L_{\mathrm{disc}}\big].
\end{aligned}
\label{eq:memory-training}
\end{equation}
Reconstruction encourages recoverable historical content. Alignment constrains original and compressed histories to yield consistent next-token predictions for the same subsequent context. Contrast matches compressed representations to their source histories and distinguishes them from other histories. Here, $i$ and $t$ index trajectories and training segments. $\omega_{i,t}$, $\beta_i$, and $\gamma_i$ are loss weights for trajectory weighting and reconstruction modulation.

For skill navigation, we jointly fine-tune the pretrained encoder and navigator to select next-step skills. At each decision point of a training trajectory, the latent representation, together with the explicit record, should rank first the skills used to support the next action:
\begin{equation}
\mathcal L_{\mathrm{nav}}=-\mathbb E_{i,t}\Big[\sum_{k\in\mathcal K} y_{i,t}(k)\log p_{\theta,\phi}\big(k\mid Z_{i,t},B_{i,t}\big)\Big],
\label{eq:nav-training}
\end{equation}
where $y_{i,t}$ distributes unit mass over the skills used at the next step and $\phi$ parameterizes the navigator. Gradients from this objective update both the encoder parameters $\theta$ and navigator parameters $\phi$, refining the pretrained state for subsequent skill decisions. Appendix~\ref{app:method-details} provides training details.

\subsection{State-Grounded Skill Decisions}
\label{sec:state-retrieval}

\textsc{JitSkill} uses the same task state to coordinate skill retrieval and execution. It retrieves reusable procedural knowledge that can advance the remaining task requirements, and provides the same state to the executor so that skill execution remains aligned with global task progress. The following components formalize these coupled decisions.

\paragraph{Progress-conditioned skill retrieval.}
As execution unfolds, new skill requirements emerge from intermediate observations and execution feedback. Rather than treating retrieval as an isolated match to local context, JitSkill retrieves skills that advance the remaining task requirements under the tracked task state. At each execution step, the navigator ranks candidate skills by their consistency with the tracked progress, then selects the currently needed skill subset $\mathcal K_t$ from the skill library $\mathcal K$. This allows the retrieved skill set to evolve with newly emerging needs while avoiding repetition of completed work, invalidated paths, and conflicts with prior execution. Appendix~\ref{app:method-details} provides the complete retrieval formulation and implementation details.

\paragraph{State-Aware Skill Execution.}
\label{sec:closed-loop-execution}
Retrieval must be coupled with skill execution: even a skill selected based on the task state can be executed inconsistently if the executor does not receive the state used to select it. JitSkill therefore supplies the executor with the same two-part state used for navigation.

We keep the executor frozen and train only the lightweight projection $P_\psi$, which maps the latent slots to state tokens $Z'_t=P_\psi(Z_t)$ that the executor can read. The executor takes as input the current instruction, observation, and recent context, together with $B_t$, $Z'_t$, and the selected skills $\mathcal K_t$, and generates the next action. For decision point $t$ of training trajectory $i$, we use the recorded next action $a_{i,t}$ as supervision and optimize the projection by:
\begin{equation}
\begin{aligned}
\mathcal L_{\mathrm{proj}}(\psi)
&=-\mathbb E_{i,t}\big[\log p_{\eta^\star}\bigl(a_{i,t}\mid u_i,o_{i,t},\\
&\qquad H_{i,t},B_{i,t},Z'_{i,t},\mathcal K_{i,t}\bigr)\big],
\end{aligned}
\label{eq:executor-training}
\end{equation}
where $\eta^\star$ denotes the fixed executor parameters, so gradients update only the projection parameters $\psi$. This objective increases the probability of the recorded action by learning state tokens that the frozen executor can use. The resulting action $a_t$ and feedback $o_{t+1}$ update $S_t$, so the next retrieval decision conditions on the observed outcome of skill execution and the updated task progress.

\section{Experiments}
\label{sec:experiments}

In this section, we conduct comprehensive experiments to answer the following research questions:
\textbf{RQ1 (Effectiveness).} How effectively does \textsc{JitSkill} support long-horizon task execution compared with existing skill-augmentation and agent-execution methods?
\textbf{RQ2 (Ablation).} How do real-time skill retrieval, progress-conditioned skill execution, and explicit and latent representations contribute to task completion?
\textbf{RQ3 (Model Studies).} How do skill injection settings affect task performance, and how efficiently does \textsc{JitSkill} complete tasks?
Further analyses appear in Appendices~\ref{app:training-curves} and~\ref{app:sensitivity}.

\begin{table*}[!t]
\centering
\caption{\textbf{Skill Retrieval Performance (\%).} We report task success rate (SR, \%) across SkillsBench, ALFWorld, and $\tau^2$-Bench. \textbf{Bold} and \underline{underlined} values mark \textbf{the best} and \underline{second-best} results. Green marks \textcolor{green!45!black}{relative improvements} in task success rate over the second-best.}
\label{tab:ts-main}
\setlength{\tabcolsep}{1.5pt}
\renewcommand{\arraystretch}{1.12}
\scriptsize
\resizebox{\textwidth}{!}{%
\begin{tabular}{>{\raggedright\arraybackslash}m{1.95cm}*{3}{>{\centering\arraybackslash}m{0.92cm}}>{\centering\arraybackslash}m{1.15cm}!{\vrule width 0.6pt}*{6}{>{\centering\arraybackslash}m{0.92cm}}>{\centering\arraybackslash}m{1.15cm}!{\vrule width 0.6pt}*{3}{>{\centering\arraybackslash}m{0.92cm}}>{\centering\arraybackslash}m{1.15cm}}
\toprule
\multirow{2}{*}{Method} & \multicolumn{4}{c!{\vrule width 0.6pt}}{\textbf{SkillsBench}} & \multicolumn{7}{c!{\vrule width 0.6pt}}{\textbf{ALFWorld}} & \multicolumn{4}{c}{\textbf{$\tau^2$-Bench}} \\
\cmidrule(lr){2-5}\cmidrule(lr){6-12}\cmidrule(lr){13-16}
& Core & Extended & Extreme & SR & Pick & Look & Clean & Heat & Cool & Pick2 & SR & Airline & Retail & Telecom & SR \\
\midrule
\multicolumn{16}{c}{\textit{Direct Execution (No Skills)}} \\
\midrule
Qwen3-8B & 0.00 & 0.00 & 3.57 & 1.15 & 71.43 & 30.77 & 29.63 & 0.00 & 0.00 & 8.33 & 27.86 & 15.00 & 2.50 & 1.67 & 6.39 \\
Qwen3.8-27B & 0.00 & 8.18 & 3.57 & 6.13 & 88.57 & 84.62 & 85.19 & 6.25 & 52.00 & 75.00 & 69.29 & 38.33 & 56.67 & 20.83 & 38.61 \\
\midrule
\multicolumn{16}{c}{\textit{Pre-Execution Skill Retrieval}} \\
\midrule
SkillTrace & \underline{11.11} & 13.84 & 8.33 & 11.88 & \underline{98.10} & 76.92 & 74.07 & 58.33 & 70.67 & 93.06 & 81.19 & \underline{48.33} & \underline{61.67} & 33.33 & 47.78 \\
SoftSkill & \textbf{16.67} & 5.66 & 7.14 & 6.90 & 89.52 & 79.49 & \underline{87.65} & 72.92 & 54.67 & 77.78 & 78.10 & 36.67 & 50.83 & 31.67 & 39.72 \\
CodeSkill & \textbf{16.67} & \underline{15.09} & 7.14 & 12.64 & 97.14 & \textbf{97.44} & 80.25 & \textbf{89.58} & 73.33 & 86.11 & 86.90 & 46.67 & 58.33 & 32.50 & 45.83 \\
SkillZip & \underline{11.11} & 6.29 & 4.76 & 6.13 & \textbf{99.05} & 79.49 & 82.72 & \underline{87.50} & 76.00 & 83.33 & 85.95 & \textbf{51.67} & 59.17 & 34.17 & \underline{48.34} \\
StructAgent & \textbf{16.67} & 10.06 & 8.33 & 9.96 & 95.24 & 53.85 & 80.25 & 81.25 & 52.00 & 87.50 & 77.86 & 38.33 & 44.17 & 4.17 & 28.89 \\
FlowState & 5.56 & 9.43 & 7.14 & 8.43 & \underline{98.10} & 89.74 & 83.95 & 72.92 & 69.33 & \textbf{95.83} & 86.19 & 45.00 & 57.50 & 20.00 & 40.83 \\
\midrule
\multicolumn{16}{c}{\textit{Execution-Time Skill Retrieval}} \\
\midrule
SGDR & \textbf{16.67} & 11.95 & 8.33 & 11.11 & 96.19 & 87.18 & 83.95 & 83.33 & 81.33 & 86.11 & 87.14 & 38.33 & \textbf{64.17} & \underline{35.00} & 45.83 \\
WebXSkill & \textbf{16.67} & 14.47 & \underline{9.52} & \underline{13.03} & \textbf{99.05} & 87.18 & 82.72 & 85.42 & 76.00 & 86.11 & 86.90 & \textbf{51.67} & \underline{61.67} & 31.67 & 48.33 \\
CUA-Skill & \textbf{16.67} & 11.95 & \underline{9.52} & 11.49 & 97.14 & 89.74 & 86.42 & 83.33 & \underline{86.67} & \underline{94.44} & \underline{90.48} & 45.00 & 57.50 & 32.50 & 45.00 \\
\midrule
\rowcolor{blue!8}
\textsc{JitSkill} (Ours) & \textbf{16.67} & \textbf{18.24} & \textbf{10.71} & \rewardgain{\textbf{15.71}}{+20.6\%} & \underline{98.10} & \underline{94.87} & \textbf{90.12} & 83.33 & \textbf{89.33} & \underline{94.44} & \rewardgain{\textbf{92.38}}{+2.1\%} & 45.00 & \textbf{64.17} & \textbf{38.33} & \rewardgain{\textbf{49.17}}{+1.7\%} \\
\bottomrule
\end{tabular}
}
\end{table*}

\subsection{Experimental Settings}

\paragraph{Datasets.} We evaluate on SkillsBench~\citep{li2026skillsbench}, ALFWorld~\citep{shridhar2021alfworld}, and $\tau^2$-Bench~\citep{barres2025tau2}, covering containerized, interactive household, and conversational tool-use tasks, respectively.

\paragraph{Baselines.} We compare three groups: (i) direct execution without skills, including Qwen3-8B and Qwen3.8-27B; (ii) pre-execution skill retrieval, including SkillTrace~\citep{yao2026skilltrace}, SoftSkill~\citep{tao2026softskill}, CodeSkill~\citep{li2026codeskill}, SkillZip~\citep{tan2026skillzip}, StructAgent~\citep{wu2026structagent}, and FlowState~\citep{li2026flowstate}; and (iii) execution-time skill retrieval, including SGDR~\citep{li2026sgdr}, WebXSkill~\citep{wang2026webxskill}, and CUA-Skill~\citep{chen2026cuaskill}.

\paragraph{Evaluation Metrics.} Following the official evaluators of the three benchmarks~\citep{li2026skillsbench,shridhar2021alfworld,barres2025tau2}, we report task success rate (SR) and mean environment interaction steps to assess task completion and interaction efficiency.

\paragraph{Implementation Details.} Methods are evaluated under shared execution and skill-access settings. Hyperparameters are selected on training data only, and results are averaged over three runs. Dataset statistics, baseline adaptations, evaluation protocols, and implementation configurations are detailed in Appendices~\ref{app:bench-stats}, \ref{app:baselines}, and~\ref{app:implementation}.

\subsection{Main Results (RQ1)}

To address RQ1, we comprehensively evaluate \textsc{JitSkill} against representative baselines across three long-horizon benchmarks under shared settings. We summarize the main observations as follows:

\newcommand{\JitSkillStudyFigures}{%
\ifnum\value{page}<7\relax
  \afterpage{\JitSkillStudyFigures}%
\else
\begin{figure*}[!t]
\centering
\begin{minipage}[t]{0.48\textwidth}
\centering
\includegraphics[width=\linewidth,height=1.35in,keepaspectratio]{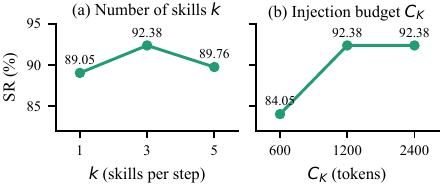}
\caption{Parameter sensitivity on ALFWorld.}
\label{fig:sensitivity}
\end{minipage}\hfill
\begin{minipage}[t]{0.50\textwidth}
\centering
\includegraphics[width=\linewidth,height=1.35in,keepaspectratio]{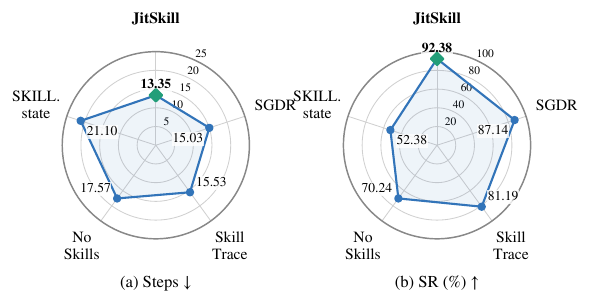}
\caption{Interaction efficiency on ALFWorld.}
\label{fig:efficiency}
\end{minipage}
\end{figure*}
\fi
}
\JitSkillStudyFigures

\paragraph{Obs.~1: \textsc{JitSkill} achieves the highest overall success rates across three long-horizon benchmarks.} Under shared settings, \textsc{JitSkill} achieves the highest overall success rates on SkillsBench ($15.71\%$), ALFWorld ($92.38\%$), and $\tau^2$-Bench ($49.17\%$). Notably, it improves over the strongest baseline on SkillsBench by $20.6\%$ relatively. Compared with direct execution using the same executor, it increases success rates by $9.6$, $23.1$, and $10.6$ percentage points, respectively. These results demonstrate the effectiveness of \textsc{JitSkill} across diverse task settings, including programmatic operations, embodied interaction, and conversational tool use. With the executor held fixed, these improvements support the effectiveness of task-state tracking in jointly guiding on-demand skill retrieval and execution throughout the task.

\paragraph{Obs.~2: State-grounded just-in-time retrieval improves on both pre-execution and execution-time retrieval baselines.} \textsc{JitSkill} exceeds the strongest methods from both retrieval groups in overall success rate on all three benchmarks. On SkillsBench, it raises success rate from $12.64\%$ for the strongest pre-execution method and $13.03\%$ for the strongest execution-time method to $15.71\%$. Its advantage also extends to the Extended and Extreme subsets, where it achieves the highest success rates while matching the best result on Core. This comparison suggests that retrieval timing alone does not account for the full benefit: updating skills during execution offers further value when skill decisions also consider prior outcomes and remaining requirements. The results are consistent with \textsc{JitSkill}'s design, which uses an evolving task state to connect emerging skill needs with accumulated progress and to guide subsequent skill execution.

\subsection{Ablation Studies (RQ2)}
To isolate the contribution of each component in \textsc{JitSkill}, we ablate just-in-time skill retrieval, progress-conditioned skill execution, the explicit progress record, and the latent progress representation on AppWorld and ALFWorld. All variants share the executor, skill library, training data, injection budget, and recent interaction context; only the indicated mechanism is changed. \emph{w/o JIT Retrieval} selects a fixed skill set from the initial instruction, whereas \emph{w/o Progress-Conditioned Execution} retains online retrieval but withholds both progress representations from the executor. The two representation variants remove the corresponding state from both retrieval and execution.
\begin{figure}[H]
\centering
\includegraphics[width=\columnwidth]{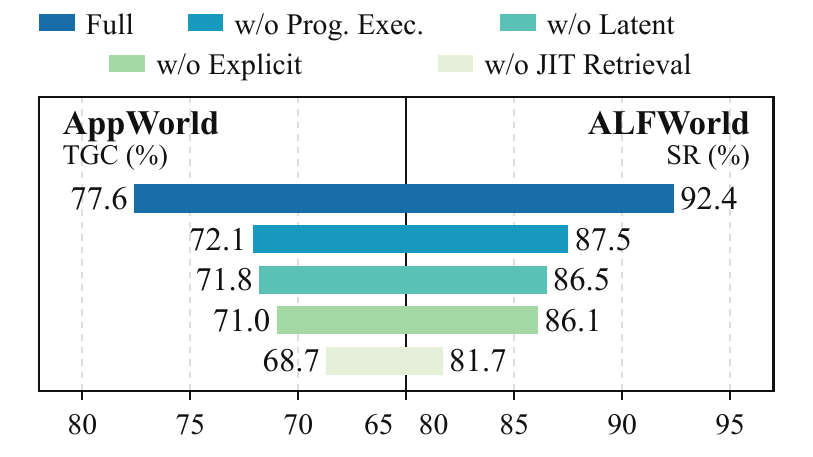}
\caption{\textbf{Module ablation of \textsc{JitSkill}.} }
\label{fig:module-ablation}
\end{figure}

\paragraph{Obs.~3: Just-in-time retrieval and progress-conditioned execution are both necessary.} Figure~\ref{fig:module-ablation} shows that fixing the initial skill set causes the largest decline: 8.9 TGC points on AppWorld and 10.7 SR points on ALFWorld. Withholding progress from the executor also lowers completion on both benchmarks despite retaining online retrieval. These results indicate that skill retrieval must respond to emerging needs, while execution must use accumulated progress to apply the retrieved skills under the current task conditions.

\paragraph{Obs.~4: Explicit and latent progress representations are complementary.} Removing either representation lowers completion on both benchmarks, with declines of 5.8--6.6 points on AppWorld and 5.9--6.3 points on ALFWorld. The explicit record preserves precise execution facts, whereas the latent state captures dependencies across interactions. Their joint use provides evidence for skill retrieval and execution grounded in task progress.

\subsection{Model Studies (RQ3)}

To answer RQ3, we study skill injection settings and compare the interaction efficiency of \textsc{JitSkill} and representative baselines on ALFWorld.

\paragraph{Obs.~5: Moderate skill counts and injection budgets yield the best performance.} Figure~\ref{fig:sensitivity} shows that increasing the skill count or injection budget does not always improve task completion. Under a fixed budget, injecting three skills outperforms injecting one or five, indicating that performance does not simply increase with the number of skills. Increasing the budget from 600 to 1,200 tokens improves success rate by 8.33 percentage points, while further increases yield no additional gains.

\paragraph{Obs.~6: \textsc{JitSkill} achieves higher task success with fewer environment interactions.} As shown in Figure~\ref{fig:efficiency}, \textsc{JitSkill} achieves the highest success rate with the fewest mean interaction steps. Compared with SGDR, it reduces mean steps from 15.03 to 13.35, indicating that its gains do not rely on more interaction attempts. This result is consistent with our motivation: continuously grounding skill acquisition and use in task progress helps advance tasks with fewer interactions. Additional runtime cost analysis is provided in Appendix~\ref{app:efficiency-alfworld}.

\section{Related Work}
\label{sec:related_work}

\noindent\textbf{Skill Retrieval.} Early work on skill retrieval focuses on pre-execution selection. SkillFlow~\citep{li2026skillflow} employs multi-stage retrieval and filtering, and SkillRouter~\citep{zheng2026skillrouter} further incorporates procedural information from skill bodies. Beyond individual skill matching, GoS~\citep{liu2026graphofskills}, GoSkills~\citep{zeng2026goskills}, and CaSKG~\citep{li2026caskg} model skill dependencies, groupings, and calibrated relations. SkillAligner~\citep{li2026skillaligner} adapts retrieved skills to task requirements and interfaces. However, pre-execution retrieval may miss procedural needs that emerge during interaction. SGDR~\citep{li2026sgdr} and WebXSkill~\citep{wang2026webxskill} dynamically retrieve skills based on the current execution context. CUA-Skill~\citep{chen2026cuaskill} further incorporates execution history into skill selection and configuration. Nevertheless, local relevance alone does not establish whether a skill remains necessary or consistent with accumulated task progress.

\noindent\textbf{State Management.} Task-state management provides another line of research on long-horizon agent execution. Existing methods address workflow control~\citep{wu2024stateflow}, progress verification~\citep{wu2026structagent}, and state-aware tool use~\citep{yang2026nod,guo2026oodatool}. Other studies focus on retaining information across interactions. FlowState~\citep{li2026flowstate} preserves execution evidence for later reuse, whereas SKILL.state~\citep{badhe2026skillstate} maintains structured state to support the execution of selected skills. Further related work appears in Appendix~\ref{app:related-work}.

\section{Conclusion}

In this paper, we formulate \emph{just-in-time skill retrieval} for long-horizon agents, acquiring skills on demand as task requirements emerge during execution. Fixed pre-execution retrieval can miss later needs, while local-context retrieval can overlook accumulated task progress and lead to execution drift. We propose \textsc{JitSkill}, which maintains complementary explicit and latent task states to make skill selection task-progress-aware rather than an independent local match, jointly guiding skill retrieval and the subsequent use of retrieved skills. 
Experiments across three long-horizon benchmarks show \textsc{JitSkill} consistently outperforms baselines, with a $20.6\%$ relative improvement over WebX-Skill on SkillsBench.
These results underscore the value of tracking evolving task progress for coherent long-horizon skill retrieval and use.

\section*{Limitations}
\textsc{JitSkill} assumes an execution interface that exposes step-wise observations and supports repeated skill retrieval during task execution. Its effectiveness may be limited when feedback is sparse or delayed, when the available skill library does not contain the capabilities required by the task, or when the maintained execution state cannot capture the information needed to distinguish competing skills. The current evaluation focuses on API-based and containerized agent environments; extending the framework to settings with richer perceptual observations, unreliable tools, or human-in-the-loop interaction remains an important direction for future work.

\bibliography{iclr2026_conference}

\begin{thebibliography}{41}
\providecommand{\natexlab}[1]{#1}

\bibitem[{Badhe et~al.(2026)Badhe, Tiwari, and Chung}]{badhe2026skillstate}
Sanket Badhe, Priyanka Tiwari, and Jonghyun Chung. 2026.
\newblock \href {https://arxiv.org/abs/2608.26263} {{SKILL.state}: Scalable
  long-horizon agent skills}.
\newblock \emph{Preprint}, arXiv:2608.26263.

\bibitem[{Bai et~al.(2026)Bai, Wan, Zhou, Yu, You, and Tsang}]{bai2026skilldag}
Tong Bai, Zhenglin Wan, Pengfei Zhou, Xingrui Yu, Yang You, and Ivor~W Tsang.
  2026.
\newblock Skilldag: Self-evolving typed skill graphs for llm skill selection at
  scale.
\newblock \emph{arXiv preprint arXiv:2606.03056}.

\bibitem[{Barres et~al.(2025)Barres, Dong, Ray, Si, and
  Narasimhan}]{barres2025tau2}
Victor Barres, Honghua Dong, Soham Ray, Xujie Si, and Karthik Narasimhan. 2025.
\newblock \href {https://arxiv.org/abs/2506.07982} {{$\tau^2$-Bench}:
  Evaluating conversational agents in a dual-control environment}.
\newblock \emph{Preprint}, arXiv:2506.07982.

\bibitem[{Chen et~al.(2026)Chen, Li, Solodko, Wang, Jiang, Cui, Hao, Ko,
  Abdali, Xu, Zheng, Fan, Cameron, Wagle, and Koishida}]{chen2026cuaskill}
Tianyi Chen, Yinheng Li, Michael Solodko, Sen Wang, Nan Jiang, Tingyuan Cui,
  Junheng Hao, Jongwoo Ko, Sara Abdali, Leon Xu, Suzhen Zheng, Hao Fan,
  Pashmina Cameron, Justin Wagle, and Kazuhito Koishida. 2026.
\newblock \href {https://arxiv.org/abs/2601.21123} {{CUA-Skill}: Develop skills
  for computer using agent}.
\newblock \emph{Preprint}, arXiv:2601.21123.

\bibitem[{Feng et~al.(2026)Feng, Xiang, Yang, Ma, Chen, Zhang, Huang, Wu, Liu,
  Wang et~al.}]{feng2026graph}
Yuyuan Feng, Zhishang Xiang, Chaobin Yang, Qichao Ma, Zerui Chen, Yujing Zhang,
  Ke~Huang, Chuanjie Wu, Zhaoxu Liu, Yili Wang, et~al. 2026.
\newblock Graph engineering in the era of llm agents: From individual
  intelligence to system intelligence.
\newblock \emph{arXiv preprint arXiv:2608.21156}.

\bibitem[{Guo et~al.(2026)Guo, Huang, Wu, Zhong, Chen, Tang, Long, and
  Hu}]{guo2026oodatool}
Rongfeng Guo, Yinxuan Huang, Yusen Wu, Maoqing Zhong, Yunlu Chen, Meng Tang,
  Teng Long, and Vincent~Tao Hu. 2026.
\newblock \href {https://arxiv.org/abs/2608.24368} {From state to action:
  {OODA-Tool} for reliable multi-turn tool use}.
\newblock \emph{Preprint}, arXiv:2608.24368.

\bibitem[{Huang et~al.(2026)Huang, Huang, Choubey, Zhou, Chen, May, and
  Wu}]{huang2026gta}
Tenghao Huang, Kung-Hsiang Huang, Prafulla~Kumar Choubey, Yilun Zhou, Muhao
  Chen, Jonathan May, and Chien-Sheng Wu. 2026.
\newblock \href {https://arxiv.org/abs/2605.29218} {Gta: Generating
  long-horizon tasks for web agents at scale}.
\newblock \emph{Preprint}, arXiv:2605.29218.

\bibitem[{Jang et~al.(2026)Jang, Koh, Fried, and
  Salakhutdinov}]{jang2026odysseys}
Lawrence~Keunho Jang, Jing~Yu Koh, Daniel Fried, and Ruslan Salakhutdinov.
  2026.
\newblock \href {https://arxiv.org/abs/2604.24964} {Odysseys: Benchmarking web
  agents on realistic long horizon tasks}.
\newblock \emph{Preprint}, arXiv:2604.24964.

\bibitem[{Kang et~al.(2026)Kang, Cho, and Kim}]{kang2026skillret}
Ryangkyung Kang, Hongcheol Cho, and Youngeun Kim. 2026.
\newblock \href {https://arxiv.org/abs/2605.05726} {Skillret: A large-scale
  benchmark for skill retrieval in llm agents}.
\newblock \emph{Preprint}, arXiv:2605.05726.

\bibitem[{Li et~al.(2025)Li, Tagkopoulos, and Tagkopoulos}]{li2026skillflow}
Fangzhou Li, Pagkratios Tagkopoulos, and Ilias Tagkopoulos. 2025.
\newblock \href {https://arxiv.org/abs/2504.06188} {Skillflow: Scalable and
  efficient agent skill retrieval system}.
\newblock \emph{Preprint}, arXiv:2504.06188.

\bibitem[{Li et~al.(2026{\natexlab{a}})Li, Deng, Wang, Huang, Shi, Tan, Lu, and
  Liu}]{li2026sgdr}
Jiaxi Li, Ke~Deng, Yun Wang, Jingyuan Huang, Yucheng Shi, Qiaoyu Tan, Jin Lu,
  and Ninghao Liu. 2026{\natexlab{a}}.
\newblock \href {https://arxiv.org/abs/2606.04391} {Online skill learning for
  web agents via state-grounded dynamic retrieval}.
\newblock \emph{Preprint}, arXiv:2606.04391.

\bibitem[{Li et~al.(2026{\natexlab{b}})Li, Li, Wang, Li, Xu, Zhang, Wu, and
  Xu}]{li2026flowstate}
Minghao Li, Bangyan Li, Zifan Wang, Yulong Li, Hu~Xu, Gan Zhang, Jingtong Wu,
  and Wenqiang Xu. 2026{\natexlab{b}}.
\newblock \href {https://arxiv.org/abs/2609.34565} {{FlowState}: Execution
  state as memory for long-horizon {LLM} agents}.
\newblock \emph{Preprint}, arXiv:2609.34565.

\bibitem[{Li et~al.(2026{\natexlab{c}})Li, He, Bao, Yang, Chen, Yu, Liang, Su,
  Zhang, and Zhang}]{li2026skillaligner}
Qinfeng Li, Dalin He, Yuntai Bao, Ying Yang, Ruoxi Chen, Xinyan Yu, Lizhou
  Liang, Ge~Su, Wenqi Zhang, and Xuhong Zhang. 2026{\natexlab{c}}.
\newblock \href {https://arxiv.org/abs/2608.06880} {Skillaligner: Treating
  retrieved skills as adaptable drafts at execution time}.
\newblock \emph{Preprint}, arXiv:2608.06880.

\bibitem[{Li et~al.(2026{\natexlab{d}})Li, Liu, Chen, You, Di, He, Zheng
  et~al.}]{li2026skillsbench}
Xiangyi Li, Yimin Liu, Wenbo Chen, Bingran You, Zonglin Di, Yifeng He, Shenghan
  Zheng, et~al. 2026{\natexlab{d}}.
\newblock \href {https://arxiv.org/abs/2602.12670} {Skillsbench: Benchmarking
  how well agent skills work across diverse tasks}.
\newblock \emph{Preprint}, arXiv:2602.12670.

\bibitem[{Li et~al.(2026{\natexlab{e}})Li, Zhang, Zhang, Liu, and
  Liu}]{li2026codeskill}
Yanzhou Li, Yiran Zhang, Xiaoyu Zhang, Xiaoxia Liu, and Yang Liu.
  2026{\natexlab{e}}.
\newblock \href {https://arxiv.org/abs/2605.25430} {{CodeSkill}: Learning
  self-evolving skills for coding agents}.
\newblock \emph{Preprint}, arXiv:2605.25430.

\bibitem[{Li et~al.(2026{\natexlab{f}})Li, Gao, Ding, Chen, Wu, and
  Chang}]{li2026caskg}
Zhiyuan Li, Linyuan Gao, Xuechun Ding, Hongwei Chen, Yuan Wu, and Yi~Chang.
  2026{\natexlab{f}}.
\newblock \href {https://arxiv.org/abs/2608.25500} {Caskg:
  Counterfactual-causal skill graphs for scalable agent skill retrieval}.
\newblock \emph{Preprint}, arXiv:2608.25500.

\bibitem[{Liu et~al.(2026)Liu, Li, Du, Wu, Gui, Kuang, and
  Sun}]{liu2026graphofskills}
Dawei Liu, Zongxia Li, Hongyang Du, Xiyang Wu, Shihang Gui, Yongbei Kuang, and
  Lichao Sun. 2026.
\newblock \href {https://arxiv.org/abs/2604.05333} {Graph-of-skills:
  Dependency-aware structural retrieval for massive agent skills}.
\newblock \emph{Preprint}, arXiv:2604.05333.

\bibitem[{Shridhar et~al.(2021)Shridhar, Yuan, C{\^o}t{\'e}, Bisk, Trischler,
  and Hausknecht}]{shridhar2021alfworld}
Mohit Shridhar, Xingdi Yuan, Marc-Alexandre C{\^o}t{\'e}, Yonatan Bisk, Adam
  Trischler, and Matthew Hausknecht. 2021.
\newblock \href {https://openreview.net/forum?id=0IOX0YjIe} {{ALFWorld}:
  Aligning text and embodied environments for interactive learning}.
\newblock In \emph{International Conference on Learning Representations
  (ICLR)}.

\bibitem[{Su et~al.(2026)Su, Long, Ai, He, Tang, Wang, Tu, Wang, and
  Liu}]{su2026sra}
Weihang Su, Jianming Long, Qingyao Ai, Qiaozhi He, Yichen Tang, Changyue Wang,
  Yiteng Tu, Yingbo Wang, and Yiqun Liu. 2026.
\newblock \href {https://arxiv.org/abs/2604.24594} {Skill retrieval
  augmentation for agentic ai}.
\newblock \emph{Preprint}, arXiv:2604.24594.

\bibitem[{Tan et~al.(2026)Tan, Wang, Liu, Xu, Yuan, Zhu, and
  Zhang}]{tan2026skillzip}
Xingyu Tan, Xiaoyang Wang, Qing Liu, Xiwei Xu, Xin Yuan, Liming Zhu, and Wenjie
  Zhang. 2026.
\newblock \href {https://arxiv.org/abs/2608.05604} {{SkillZip}:
  Contract-preserving graph compression for scalable agent skill libraries}.
\newblock \emph{Preprint}, arXiv:2608.05604.

\bibitem[{Tao et~al.(2026)Tao, Teng, Fu, Liu, Chen, Zhao, Zhang, Liu, and
  Kong}]{tao2026softskill}
Xijia Tao, Yihua Teng, Xinyu Fu, Ziru Liu, Kecheng Chen, Yuzhi Zhao, Suiyun
  Zhang, Rui Liu, and Lingpeng Kong. 2026.
\newblock \href {https://arxiv.org/abs/2606.20333} {{SoftSkill}: Behavioral
  compression for contextual adaptation}.
\newblock \emph{Preprint}, arXiv:2606.20333.

\bibitem[{Wang et~al.(2023)Wang, Xie, Jiang, Mandlekar, Xiao, Zhu, Fan, and
  Anandkumar}]{wang2023voyager}
Guanzhi Wang, Yuqi Xie, Yunfan Jiang, Ajay Mandlekar, Chaowei Xiao, Yuke Zhu,
  Linxi Fan, and Anima Anandkumar. 2023.
\newblock \href {https://arxiv.org/abs/2305.16291} {Voyager: An open-ended
  embodied agent with large language models}.
\newblock \emph{Preprint}, arXiv:2305.16291.

\bibitem[{Wang et~al.(2026)Wang, Wu, Zhang et~al.}]{wang2026webxskill}
Zhaoyang Wang, Qianhui Wu, Xuchao Zhang, et~al. 2026.
\newblock \href {https://arxiv.org/abs/2604.13318} {{WebXSkill}: Skill learning
  for autonomous web agents}.
\newblock \emph{Preprint}, arXiv:2604.13318.

\bibitem[{Wu et~al.(2026{\natexlab{a}})Wu, Zhu, Zhou, Salvi, Song, and
  Huang}]{wu2026structagent}
Wenyi Wu, Sibo Zhu, Kun Zhou, Aayush Salvi, Zixuan Song, and Biwei Huang.
  2026{\natexlab{a}}.
\newblock \href {https://arxiv.org/abs/2607.11388} {{StructAgent}: Harness
  long-horizon digital agents with unified causal structure}.
\newblock \emph{Preprint}, arXiv:2607.11388.

\bibitem[{Wu et~al.(2026{\natexlab{b}})Wu, Li, Shi, Duffy, Marques, Olson,
  Zhou, and Manocha}]{wu2026cosplay}
Xiyang Wu, Zongxia Li, Guangyao Shi, Alexander Duffy, Tyler Marques,
  Matthew~Lyle Olson, Tianyi Zhou, and Dinesh Manocha. 2026{\natexlab{b}}.
\newblock \href {https://arxiv.org/abs/2604.20987} {Co-evolving llm decision
  and skill bank agents for long-horizon tasks}.
\newblock \emph{Preprint}, arXiv:2604.20987.

\bibitem[{Wu et~al.(2024)Wu, Yue, Zhang, Wang, and Wu}]{wu2024stateflow}
Yiran Wu, Tianwei Yue, Shaokun Zhang, Chi Wang, and Qingyun Wu. 2024.
\newblock \href {https://arxiv.org/abs/2403.11322} {{StateFlow}: Enhancing
  {LLM} task-solving through state-driven workflows}.
\newblock \emph{Preprint}, arXiv:2403.11322.

\bibitem[{Xia et~al.(2026)Xia, Chen, Wang, Liu, Zeng, Wang, Han, Zhou, Zhao,
  Chen, Zheng, Xie, and Yao}]{xia2026skillrl}
Peng Xia, Jianwen Chen, Hanyang Wang, Jiaqi Liu, Kaide Zeng, Yu~Wang, Siwei
  Han, Yiyang Zhou, Xujiang Zhao, Haifeng Chen, Zeyu Zheng, Cihang Xie, and
  Huaxiu Yao. 2026.
\newblock \href {https://arxiv.org/abs/2602.08234} {Skillrl: Evolving agents
  via recursive skill-augmented reinforcement learning}.
\newblock \emph{Preprint}, arXiv:2602.08234.

\bibitem[{Xiang et~al.(2026)Xiang, Yang, Chen, Wei, Tang, Teng, Peng, Li,
  Huang, He et~al.}]{xiang2026systematic}
Zhishang Xiang, Chengyi Yang, Zerui Chen, Zhimin Wei, Yunbo Tang, Zongpei Teng,
  Zexi Peng, Zongxia Li, Chengsong Huang, Yicheng He, et~al. 2026.
\newblock A systematic survey of self-evolving agents: From model-centric to
  environment-driven co-evolution.
\newblock \emph{Available at SSRN 6626878}.

\bibitem[{Xu et~al.(2026)Xu, Yang, and Huang}]{xu2026hyperskill}
Ruiyao Xu, Tiankai Yang, and Wei-Chieh Huang. 2026.
\newblock Hyperskill: Self-evolving llm agents via hypergraph-structured skill
  memory.
\newblock \emph{arXiv preprint arXiv:2608.16114}.

\bibitem[{Yang et~al.(2026)Yang, Zheng, Jiang, Tang, Zhang, Wu, Chen, and
  Yu}]{yang2026nod}
Zixu Yang, Hang Zheng, Nan Jiang, Zhiyang Tang, Situo Zhang, Xiaobao Wu,
  Lu~Chen, and Kai Yu. 2026.
\newblock \href {https://arxiv.org/abs/2605.12240} {No action without a {NOD}:
  A heterogeneous multi-agent architecture for reliable service agents}.
\newblock \emph{Preprint}, arXiv:2605.12240.

\bibitem[{Yao et~al.(2023)Yao, Zhao, Yu, Du, Shafran, Narasimhan, and
  Cao}]{yao2023react}
Shunyu Yao, Jeffrey Zhao, Dian Yu, Nan Du, Izhak Shafran, Karthik Narasimhan,
  and Yuan Cao. 2023.
\newblock React: Synergizing reasoning and acting in language models.
\newblock In \emph{International Conference on Learning Representations
  (ICLR)}.

\bibitem[{Yao et~al.(2026)Yao, Wang, Chen, Zhang, He, Luo, and
  Gedeon}]{yao2026skilltrace}
Yue Yao, Shengyuan Wang, Xin Chen, Minke Zhang, Jia He, Bingjun Luo, and Tom
  Gedeon. 2026.
\newblock \href {https://arxiv.org/abs/2608.02356} {Skilltrace: Traversing a
  query-skill graph for composable llm agents}.
\newblock \emph{Preprint}, arXiv:2608.02356.

\bibitem[{Yu et~al.(2026)Yu, Zhu, Lin, Cui, Ding, and Li}]{yu2026masa}
Jianxiang Yu, Jiapeng Zhu, Bochen Lin, Qier Cui, Zichen Ding, and Xiang Li.
  2026.
\newblock \href {https://arxiv.org/abs/2605.30723} {Skill is not
  one-size-fits-all: Model-aware skill alignment for llm agents}.
\newblock \emph{Preprint}, arXiv:2605.30723.

\bibitem[{Yuan et~al.(2026)Yuan, Zhou, Xiong, Wu, Sun, Song, Cui, Wang, Wu, Li,
  Lu, Lu, Zhen et~al.}]{yuan2026osworld2}
Mengqi Yuan, Zilong Zhou, Xinzhuang Xiong, Weiming Wu, Jiayang Sun, Jiamin
  Song, Kaiqian Cui, Bowen Wang, Haoyuan Wu, Yitong Li, Dunjie Lu, Haikong Lu,
  Qi~Zhen, et~al. 2026.
\newblock \href {https://arxiv.org/abs/2606.29537} {Osworld 2.0: Benchmarking
  computer use agents on long-horizon real-world tasks}.
\newblock \emph{Preprint}, arXiv:2606.29537.

\bibitem[{Zeng et~al.(2026)Zeng, Huo, Zhang, Ye, Zhuo, Liu, Lu, Wen, and
  Tang}]{zeng2026goskills}
Kun Zeng, Yu~Huo, Siyu Zhang, Zi~Ye, Yuecheng Zhuo, Haoyue Liu, Yuquan Lu,
  Junhao Wen, and Xiaoying Tang. 2026.
\newblock \href {https://arxiv.org/abs/2605.06978} {Group of skills:
  Group-structured skill retrieval for agent skill libraries}.
\newblock \emph{Preprint}, arXiv:2605.06978.

\bibitem[{Zhang et~al.(2026)Zhang, He, Fu, Feng, Li, Wang, Wang, and
  Zhang}]{zhang2026skillpoison}
Lizhi Zhang, Xin He, Dianxuan Fu, Yuyuan Feng, Jiatong Li, Qi~Wang, Xin Wang,
  and Qinggang Zhang. 2026.
\newblock Skillpoison: Progressive skill poisoning via successful experiences.
\newblock \emph{arXiv preprint arXiv:2610.07645}.

\bibitem[{Zheng et~al.(2026)Zheng, Zhang, Ma, Yu, Zhu, Wu, Xu, Dong, Zhu,
  Huang, and Yu}]{zheng2026skillrouter}
YanZhao Zheng, ZhenTao Zhang, Chao Ma, YuanQiang Yu, JiHuai Zhu, Yong Wu,
  Tianze Xu, Baohua Dong, Hangcheng Zhu, Ruohui Huang, and Gang Yu. 2026.
\newblock \href {https://arxiv.org/abs/2603.22455} {Skillrouter: Skill routing
  for llm agents at scale}.
\newblock \emph{Preprint}, arXiv:2603.22455.

\bibitem[{Zhou et~al.(2026{\natexlab{a}})Zhou, Guo, Liu, Yu, Gong, Zhao, Chen,
  Zhang, Chen, Li et~al.}]{zhou2026memento}
Huichi Zhou, Siyuan Guo, Anjie Liu, Zhongwei Yu, Ziqin Gong, Bowen Zhao, Zhixun
  Chen, Menglong Zhang, Yihang Chen, Jinsong Li, et~al. 2026{\natexlab{a}}.
\newblock Memento-skills: Let agents design agents.
\newblock \emph{arXiv preprint arXiv:2603.18743}.

\bibitem[{Zhou et~al.(2024)Zhou, Xu, Zhu, Zhou, Lo, Sridhar, Cheng, Ou, Bisk,
  Fried, Alon, and Neubig}]{zhou2024webarena}
Shuyan Zhou, Frank~F. Xu, Hao Zhu, Xuhui Zhou, Robert Lo, Abishek Sridhar,
  Xianyi Cheng, Tianyue Ou, Yonatan Bisk, Daniel Fried, Uri Alon, and Graham
  Neubig. 2024.
\newblock \href
  {https://proceedings.iclr.cc/paper_files/paper/2024/hash/4410c0711e9154a7a2d26f9b3816d1ef-Abstract-Conference.html}
  {Webarena: A realistic web environment for building autonomous agents}.
\newblock In \emph{International Conference on Learning Representations
  (ICLR)}.

\bibitem[{Zhou et~al.(2026{\natexlab{b}})Zhou, Wang, Dai, Zhang, and
  Wang}]{zhou2026eviskill}
Yan Zhou, Yili Wang, Yiwei Dai, Qinggang Zhang, and Xin Wang.
  2026{\natexlab{b}}.
\newblock Eviskill: Grounding skill evolution in replayable evidence.
\newblock \emph{arXiv preprint arXiv:2610.05030}.

\bibitem[{Zhu et~al.(2026)Zhu, Amiri, Chen, and
  Murugesan}]{zhu2026skillmemcompactexternalskill}
Yuxuan Zhu, Mohammad~Mohammadi Amiri, Pin-Yu Chen, and Keerthiram Murugesan.
  2026.
\newblock \href {https://arxiv.org/abs/2610.37935} {Skillmem: Compact external
  skill memory for latent multi-agent reasoning}.
\newblock \emph{Preprint}, arXiv:2610.37935.

\end{thebibliography}

\clearpage
\appendix
\onecolumn
\startcontents[app]
\section*{Appendix Contents}
\printcontents[app]{}{1}{\setcounter{tocdepth}{2}}
\clearpage
\twocolumn
\renewcommand{\dbltopfraction}{0.9}
\renewcommand{\dblfloatpagefraction}{0.8}
\section{Frequently Asked Questions}
\label{app:faq}

\subsection{What are the advantages of \textsc{JitSkill}?}

Existing skill-augmented agents either fix the skill set before execution or retrieve skills from the current local signal, such as the latest observation or a summary of the current state. Neither tells the agent whether a seemingly relevant skill fills an unmet requirement, repeats completed work, or conflicts with earlier execution. \textsc{JitSkill} maintains an evolving task-progress state, consisting of an explicit progress record and a latent state, and uses the same state both to select skills and to condition the executor that applies them. As a result, the selected skills follow the stage of the task (Appendix~\ref{app:skill-usage} and Appendix~\ref{app:case-study}), and \textsc{JitSkill} achieves the best success rate on SkillsBench and ALFWorld under the same executor and skill budget (Table~\ref{tab:ts-main}).

\subsection{Why maintain both an explicit record and a latent state?}

The two representations carry complementary information. The explicit record keeps discrete, inspectable facts such as completed operations and failed attempts, whereas the latent state carries cross-step context that is difficult to enumerate as fixed textual fields. Removing either one lowers performance on both ALFWorld and AppWorld (Figure~\ref{fig:module-ablation}), so the full state is needed for the best result.

\subsection{Does \textsc{JitSkill} benefit from a larger context than the baselines?}

No. All skill-augmented methods deliver their skills at the same prompt location under the same per-step budget, and the explicit record of \textsc{JitSkill} is counted within this budget. \textsc{JitSkill} changes which skills are delivered, not how much text is delivered: it switches skills more often than step-wise retrieval while every injected block stays within the budget (Appendix~\ref{app:skill-usage}), and it completes ALFWorld games in fewer steps than the no-skill agent and SGDR (Appendix~\ref{app:efficiency-alfworld}).

\subsection{How are the baselines made comparable?}

Each baseline is re-implemented under one shared protocol: the same frozen executor, skill library, retrieval backbone, injection location, and skill budget. Each method follows the default configuration of its paper, any extra model call it requires is issued to the same executor, and every hyper-parameter is selected on official training data only (Appendix~\ref{app:baselines}).

\subsection{How sensitive is \textsc{JitSkill} to its hyper-parameters?}

\textsc{JitSkill} stays above $89\%$ on ALFWorld for every number of injected skills from $1$ to $5$, and a moderate skill budget already reaches the best success rate (Appendix~\ref{app:sensitivity}). The weights of the training objective lie on a flat optimum (Appendix~\ref{app:training-curves}).

\subsection{Does \textsc{JitSkill} generalize beyond the evaluation setting?}

Yes. Without re-tuning, \textsc{JitSkill} matches the highest success rate on unseen ALFWorld environments (Appendix~\ref{app:alfw-unseen}). With a weaker Llama-3.1-8B-Instruct executor, it achieves the highest success rate, whereas step-wise retrieval falls below the no-skill agent; the navigator remains unchanged and any executor-side interface is trained on official training trajectories only (Appendix~\ref{app:executor-transfer}).

\FloatBarrier
\section{Related Work}
\label{app:related-work}

\subsection{Pre-Execution Skill Preparation}
\label{app:skill-preparation}
Some studies treat skill augmentation as preparation before execution, acquiring relevant skills from an external library based on initial task requirements to provide procedural support for subsequent execution.
SkillFlow~\citep{li2026skillflow} uses multi-stage retrieval and filtering to obtain a compact skill set from a large library.
SkillRouter~\citep{zheng2026skillrouter} further incorporates procedural information from skill bodies, addressing the limitations of assessing relevance from names and short descriptions alone.
Since the relevance of individual skills does not guarantee that the resulting set supports complete execution, Graph-of-Skills (GoS)~\citep{liu2026graphofskills} incorporates skill dependencies into retrieval to recover prerequisites that might otherwise be missed.
GoSkills~\citep{zeng2026goskills} extends the retrieval unit to role-labeled skill groups and presents the selected skills through Start, Support, Check, and Avoid fields, helping the executor understand how they work together.
CaSKG~\citep{li2026caskg} calibrates skill relations through counterfactual probes, allowing graph-based retrieval to expand the retrieved skill set along more reliable procedural relations.
Beyond improving the acquired skill set, SkillAligner~\citep{li2026skillaligner} jointly adapts retrieved skills before execution, specializing skill fragments to task requirements and available interfaces, resolving dependencies, conflicts, and redundancy, and producing a guide for subsequent execution.
Starting from initial requirements, these methods improve the relevance, completeness, and applicability of the skills available when execution begins.
However, the capability demands of long-horizon tasks often emerge gradually through environmental feedback, while acquired resources and newly discovered constraints change the procedural support needed later.
Our work focuses on these changes during execution, continuously maintaining task progress so that skill acquisition and execution can adapt together to the remaining requirements.

\subsection{Skill Acquisition and Use During Execution}
\label{app:execution-time-skills}
As interaction environments and capability demands change during execution, some web and computer-use agents dynamically acquire skills and execute them while a task is in progress~\citep{xiang2026systematic}.
SGDR~\citep{li2026sgdr} extracts reusable subprocedures of intermediate granularity from completed trajectories, representing their purpose in natural language and providing operational support through code.
It combines the task goal with the current webpage state to retrieve skills at each step, rather than restricting skill execution to a fixed set selected at the start of the task.
WebXSkill~\citep{wang2026webxskill} identifies currently available skills from URL patterns and page elements, uses task and page context for semantic matching across websites, and provides both callable action sequences and step-level guidance to support direct execution or agent-controlled actions.
These mechanisms emphasize matching skills to the current environment.
However, local page conditions or semantic similarity do not fully capture previously completed work and established constraints, and the same page may call for different operations at different stages of a task.
Consequently, locally relevant skills may still lead an agent to repeat completed exploration or proceed before prerequisites have been satisfied.
CUA-Skill~\citep{chen2026cuaskill} represents computer operations as parameterized skills and execution graphs.
It reinitializes memory for each task and uses this memory to provide execution history for skill selection and argument configuration, helping the agent determine how to complete the current operation successfully in the current interface.
Within skill configuration and execution, the emphasis is on reliably carrying out an already selected operational skill.
Our work instead examines whether that skill still contributes to task progress, continuously maintaining task progress to jointly condition subsequent skill acquisition and execution.

\begin{figure*}[!t]
\centering
\includegraphics[width=\textwidth,trim=0 200bp 0 0,clip]{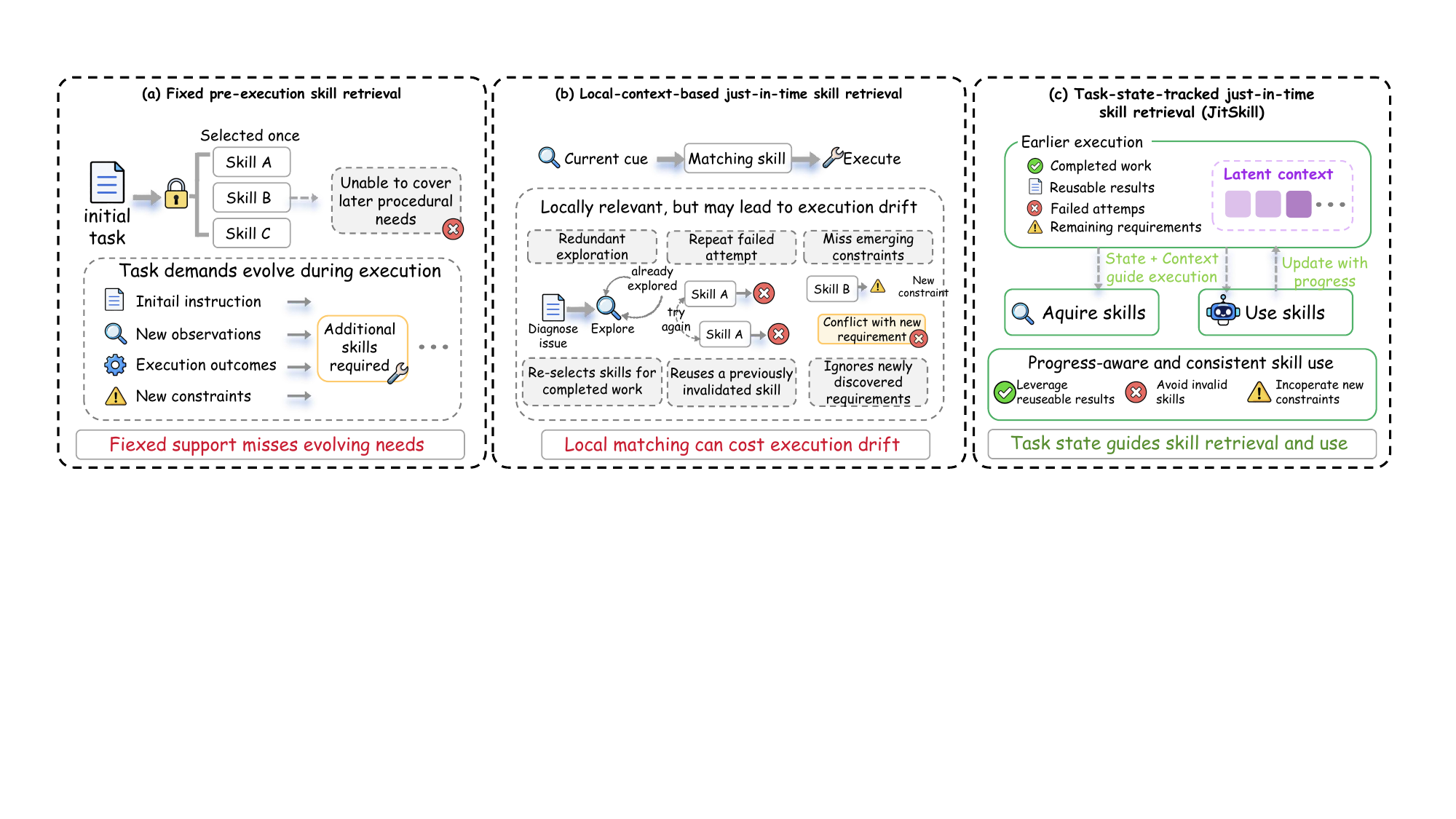}
\caption{Different skill retrieval modes. \textbf{(a)} Fixed pre-execution retrieval can miss emerging procedural needs. \textbf{(b)} Local-context-based just-in-time retrieval can overlook accumulated progress, resulting in execution drift. \textbf{(c)} \textsc{JitSkill} maintains task-progress representations to jointly guide skill retrieval and execution.}
\label{fig:appendix-skill-retrieval-modes}
\end{figure*}

\subsection{Execution Management for Long-Horizon Task}
\label{app:task-state-tracking}
Maintaining completed work, unmet requirements, and existing constraints can help agents reduce errors in assessing task progress and improve the consistency of subsequent decisions during long-horizon interaction.
To improve execution controllability, StateFlow~\citep{wu2024stateflow} organizes complex tasks into states and transitions, distinguishing workflow progression from task solving within each stage so that operations follow explicit execution stages.
Rather than primarily controlling workflow progression, StructAgent~\citep{wu2026structagent} focuses on whether an agent has sufficient evidence to conclude that a task has made progress.
It confirms state changes through verification, preventing attempted operations from being treated as completed work and supporting task-completion judgments and failure recovery.
It also introduces relevant operational experience at subgoal transitions, while emphasizing verifiable progress, error localization, and recovery.
To address failures to carry historical information into subsequent operations, NOD~\citep{yang2026nod} combines explicit task state with critical-action review, while OODA-Tool~\citep{guo2026oodatool} separates state reconstruction from action generation through staged processing.
Both aim to reduce the neglect of existing requirements and constraints in tool calls, focusing on operation correctness and consistency with task state during multi-turn interaction.
Considering the reuse of historical information in later interactions, FlowState~\citep{li2026flowstate} observes that some information becomes relevant only when subsequent subtasks or requests arise, whereas repeated context compression can discard it prematurely.
It therefore preserves execution states and associated evidence as memory reusable across requests and revisits them when new decisions require them, addressing the trade-off between persistent information retention and working-context cost.
Compared with these approaches to general agent execution, SKILL.state~\citep{badhe2026skillstate} directly addresses long-horizon execution after a skill has already been selected.
To mitigate growing interaction histories and interference from obsolete information, it replaces accumulated conversation records with an updatable structured state, allowing agents to execute the selected skill effectively in the execution environment based on the current state and latest observation.
Our work focuses on how task progress changes the need for skill retrieval.
Beyond improving operation reliability or assessing execution outcomes, task state serves as a shared basis for skill acquisition and execution, allowing completed work, remaining requirements, and existing constraints to jointly determine which skills are needed next and how the same skill should be executed at different stages of task progress.

\paragraph{From Retrieval Timing to Task-Progress Grounding.}
The preceding discussion distinguishes when skills are retrieved from how skill retrieval and execution account for task progress. Figure~\ref{fig:appendix-skill-retrieval-modes} illustrates this distinction. Fixed pre-execution retrieval can miss procedural needs revealed by later interaction. Just-in-time retrieval responds to these emerging needs, but decisions based on local context can still overlook completed work, failure conditions, and newly established constraints. \textsc{JitSkill} addresses this gap by maintaining an evolving task-progress state that jointly informs skill retrieval and execution.

A detailed organization of the related work is provided in Table~\ref{tab:baseline-design}.

\begin{table*}[t]
\centering
\caption{\textbf{Representative methods related to \textsc{JitSkill}.} Methods are grouped by when and how they construct, retrieve, or maintain the skill set used during execution.}
\label{tab:baseline-design}
\footnotesize
\setlength{\tabcolsep}{5pt}
\renewcommand{\arraystretch}{1.16}
\begin{tabularx}{\textwidth}{@{}>{\raggedright\arraybackslash}p{0.25\textwidth} >{\raggedright\arraybackslash}p{0.35\textwidth} >{\raggedright\arraybackslash}X@{}}
\toprule
\textbf{Method} & \textbf{Method description} & \textbf{Differences from ours} \\
\midrule
\rowcolor{gray!12}
\multicolumn{3}{@{}l@{}}{\strut\textbf{\(\blacktriangleright\) Pre-execution fixed skill retrieval}} \\
\midrule
SkillFlow~\citep{li2026skillflow}; SkillRouter~\citep{zheng2026skillrouter}
& Retrieve skills from initial task requirements; SkillRouter additionally uses procedural information in skill bodies.
& Determine the retrieved skill set before execution, without updating it as task progress changes. \\
\addlinespace
Graph-of-Skills (GoS)~\citep{liu2026graphofskills}; GoSkills~\citep{zeng2026goskills}; CaSKG~\citep{li2026caskg}
& Expand the retrieved skill set through dependencies, role-labeled groups, or calibrated skill relations.
& Improve the initial support set, but do not condition retrieval on evolving task state. \\
\addlinespace
SkillAligner~\citep{li2026skillaligner}
& Adapt retrieved skill fragments to task requirements and interfaces, resolving conflicts before execution.
& Improves pre-execution compatibility; later feedback does not trigger renewed support selection. \\
\addlinespace
SkillTrace~\citep{yao2026skilltrace}
& Decompose a task into subqueries, match them to skills, and add dependency-based supporting skills.
& A static skill-context baseline: it fixes the retrieved skill set before execution rather than revising it from interaction feedback. \\
\midrule
\rowcolor{gray!12}
\multicolumn{3}{@{}l@{}}{\strut\textbf{\(\blacktriangleright\) Execution-time local-signal skill retrieval}} \\
\midrule
SGDR~\citep{li2026sgdr}
& Retrieve reusable subprocedures from task-goal similarity and a summary of the current environment state.
& Uses current local state, but does not maintain a persistent representation of completed work or remaining constraints. \\
\addlinespace
WebXSkill~\citep{wang2026webxskill}
& Retrieve skills using URL patterns, page elements, and task--page semantic matching.
& Relies on static graph organization plus current-page grounding; the same page cannot be interpreted differently from task progress. \\
\addlinespace
CUA-Skill~\citep{chen2026cuaskill}
& Retrieve, rerank, and configure parameterized skills from UI context, execution history, and outcome records.
& Focuses on executing a selected operation successfully; our method also decides whether the retrieved skill remains useful for the task. \\
\midrule
\rowcolor{gray!12}
\multicolumn{3}{@{}l@{}}{\strut\textbf{\(\blacktriangleright\) Cross-step task-state maintenance}} \\
\midrule
\textsc{SKILL.state}~\citep{badhe2026skillstate}
& Replace accumulated history with a structured state, procedural specification, and latest observation.
& Maintains state for executing an already selected skill, rather than selecting the next skill. \\
\addlinespace
StructAgent~\citep{wu2026structagent}
& Verify progress using explicit evidence and support error localization and recovery.
& Focuses on reliable progress assessment, not on jointly governing skill acquisition and execution. \\
\addlinespace
FlowState~\citep{li2026flowstate}
& Preserve execution states and evidence for reuse when later subtasks require them.
& Revisits retained information when needed, but does not use progress state to control the skill lifecycle. \\
\bottomrule
\end{tabularx}
\end{table*}

\section{Method Details}
\label{app:method-details}

This appendix details the recurrent encoder, its history-compression pretraining objective, the navigation fine-tuning objective and skill scoring used in our experiments, and the training configuration.

\begin{table*}[!t]
\centering
\caption{\textbf{Training configuration of the trainable components of \textsc{JitSkill}.}}
\label{tab:method-hparams}
\footnotesize
\setlength{\tabcolsep}{3pt}
\begin{tabular*}{\textwidth}{@{\extracolsep{\fill}}llll@{}}
\toprule
Setting & Value & Setting & Value \\
\midrule
Encoder base model & Qwen3-8B (frozen) & Tokens per update (new block / encoder window) & 1{,}536 / 4{,}096 \\
Executor & Qwen3.8-27B (frozen) & Skill and record embedder & all-MiniLM-L6-v2 \\
LoRA rank / $\alpha$ / dropout & 32 / 64 / 0.05 & Epochs & 3 \\
LoRA target modules & $q,k,v,o$ projections & Learning rate (encoder / navigator) & $5\times10^{-5}$ / $10^{-3}$ \\
Latent slots $m$ & 16 & Held-out validation scenarios & 5 \\
\bottomrule
\end{tabular*}
\end{table*}

\subsection{Recurrent Encoder}

The latent representation uses $m=16$ slots. The encoder $f_\theta$ is a frozen base language model adapted with LoRA: at each step it reads the previous representation $Z_t$, the token embeddings of the newly accumulated interaction $D_{t+1}$, and $m$ learnable slots, and the outputs at the slot positions form $Z_{t+1}$. The initial representation and the learnable slots are trained together with the LoRA weights. Each update reads only the previous representation and the new block, so the per-step cost does not grow with trajectory length.

To match history integration at inference, training constructs successive history blocks along each trajectory and passes the representation from one step to the next. Forward computation preserves this recurrent dependency, while gradients are truncated across steps, which keeps the memory cost of training independent of trajectory length.

\subsection{History-Compression Pretraining}

The first training stage pretrains the encoder to retain historical content, support its subsequent use, and keep different histories distinguishable. We constrain content recovery through reconstruction, subsequent prediction through distribution alignment, and distinctions between histories through representation contrast.

\paragraph{History reconstruction.}
This objective constrains retained information by reconstructing historical fragments. Given a target fragment $y=(y_1,\ldots,y_n)$, a frozen decoder predicts each token conditioned on the compressed representation $Z$ and the ground-truth prefix $y_{<j}$:
\begin{equation}
\mathcal L_{\mathrm{rec}}
=-\frac{1}{n}\sum_{j=1}^{n}\log p_0(y_j\mid Z,y_{<j}).
\label{eq:history-reconstruction}
\end{equation}

\paragraph{Distribution alignment.}
This objective constrains compressed and original histories to yield consistent prediction distributions for the same subsequent context. Let $h$ be the history to compress and $r$ the subsequent text. The frozen teacher reads $[h;r]$, while the student reads $[Z;r]$. Their next-token distributions at the final position, $p_T$ and $p_S$, are aligned using Jensen--Shannon divergence:
\begin{equation}
\mathcal L_{\mathrm{align}}=\operatorname{JSD}(p_T\Vert p_S)/\ln 2.
\label{eq:distribution-alignment}
\end{equation}

\paragraph{Representation contrast.}
This objective preserves distinctions between histories by matching compressed representations to their source histories. Averaging $Z$ across slots and normalizing yields $v$. The positive $q^+$ is the centered mean hidden representation of the corresponding history from a frozen model; negatives come from a history-representation queue $\mathcal Q^-$:
\begin{equation}
\mathcal L_{\mathrm{disc}}
=-\log\frac{\exp(v^\top q^+/\tau)}
{\begin{gathered}
\exp(v^\top q^+/\tau)\\
+\sum_{q^-\in\mathcal Q^-}\exp(v^\top q^-/\tau)
\end{gathered}}.
\label{eq:representation-contrast}
\end{equation}
Here, $\tau$ is the temperature. Positive and negative history anchors are normalized and receive no gradients.

\paragraph{Joint weighting.}
With trajectory weighting and segment-specific reconstruction modulation, the compression objective is
\begin{equation}
\begin{aligned}
\mathcal L_{\mathrm{comp}}=\mathbb E_{i,t}\Big[w_i\Big(&
(\lambda_{\mathrm{rec}}+\lambda_{\mathrm{probe}}s_{i,t})\mathcal L_{\mathrm{rec}}\\
&+\lambda_{\mathrm{align}}\mathcal L_{\mathrm{align}}
+\lambda_{\mathrm{disc}}\mathcal L_{\mathrm{disc}}\Big)\Big].
\end{aligned}
\label{eq:memory-training-full}
\end{equation}
Here, $w_i$ is the trajectory weight, the coefficients $\lambda_{\mathrm{rec}}$, $\lambda_{\mathrm{align}}$, and $\lambda_{\mathrm{disc}}$ control the three losses, and the probe weight $s_{i,t}$ modulates reconstruction through $\lambda_{\mathrm{probe}}$. The effective weights in Equation~\ref{eq:memory-training} are
\begin{equation}
\begin{aligned}
\omega_{i,t}&=w_i(\lambda_{\mathrm{rec}}+\lambda_{\mathrm{probe}}s_{i,t}),\\
\beta_i&=w_i\lambda_{\mathrm{align}},\qquad
\gamma_i=w_i\lambda_{\mathrm{disc}}.
\end{aligned}
\end{equation}

\subsection{Navigation Objective and Skill Scoring}

The navigator $g_\phi$ maps the slot-averaged latent representation and a sentence embedding of the explicit record $B_t$ to a query vector $q_t$. Each skill $k\in\mathcal K$ is scored by the similarity between $q_t$ and the embedding of its description, which defines $p_{\theta,\phi}(k\mid Z_t,B_t)$ through a softmax over the library. At inference, this learned score is combined with the relevance of each skill to the current step, and the top-ranked skills are injected within the skill-injection budget $C_K$.

The second training stage initializes the encoder with the pretraining checkpoint and jointly fine-tunes it with the navigator on execution trajectories from the official training split. For each decision point, the supervision target is the set of skills that support the next action, determined from the operations the trajectory executes at that step. The target $y_{i,t}$ in Equation~\ref{eq:nav-training} spreads unit mass uniformly over this set, and decision points with no supporting skill are skipped. This objective jointly updates the encoder and navigator in all experiments. For benchmarks whose skill library differs from the training library, we use a scoring head that reads skill descriptions directly; it is trained on the same AppWorld training trajectories and applies to a new library without retraining.

\subsection{Executor Interface}

The 27B executor remains frozen. When the encoder's latent space differs from the executor's input embedding space, the latent slots $Z_t\in\mathbb R^{m\times d}$ are mapped to executor-space state tokens through a learned projection,
\begin{equation}
Z'_t=P_\psi(Z_t)\in\mathbb R^{m\times d_{\mathrm{exec}}},
\label{eq:latent-state-projection}
\end{equation}
where $d_{\mathrm{exec}}$ is the executor's token dimension. At each decision point, the executor receives the explicit record $B_t$, $Z'_t$, and the selected skills $\mathcal K_t$, in addition to the current instruction, observation, and recent history. We optimize only $P_\psi$ with the next-action likelihood in Equation~\ref{eq:executor-training}; the executor parameters remain fixed. This aligns the state-token interface with the same two-part task-progress state used for navigation.

\subsection{Algorithms}
\label{app:algorithms}

Algorithm~\ref{alg:inference} summarizes how \textsc{JitSkill} tracks task progress and uses it for skill acquisition and skill use at every decision point, and Algorithm~\ref{alg:training} summarizes how its trainable components are learned. Both follow the notation of Section~\ref{sec:method}.

\begin{algorithm}[!htb]
\caption{Progress-grounded skill retrieval and use}
\label{alg:inference}
\small
\SetKwInOut{Input}{Require}\SetKwInOut{Output}{Ensure}\SetCommentSty{textrm}\SetKwComment{Comment}{$\triangleright$\ }{}
\Input{instruction $u$, skill library $\mathcal K$, encoder $f_\theta$, navigator $g_\phi$, projection $P_\psi$, frozen executor $\pi_{\eta^\star}$, skill budget $C_K$, step limit $T$}
\Output{trajectory $\tau$}
$o_0\leftarrow$ initial observation; $H_0\leftarrow\varnothing$; $Z_0\leftarrow Z_{\mathrm{init}}$; $B_0\leftarrow\mathcal E(u,\varnothing)$\;
\For{$t=0,1,\ldots,T-1$}{
  \Comment{progress-conditioned skill acquisition}
  $q_t\leftarrow g_\phi(Z_t,B_t)$\;
  rank $k\in\mathcal K$ by $p_{\theta,\phi}(k\mid Z_t,B_t)$ and the current step\;
  $\mathcal K_t\leftarrow$ top-ranked skills that fit $C_K$ together with $B_t$\;
  \Comment{progress-conditioned skill use}
  $Z'_t\leftarrow P_\psi(Z_t)$\;
  $a_t\sim\pi_{\eta^\star}(\cdot\mid u,o_t,H_t,B_t,Z'_t,\mathcal K_t)$\;
  $o_{t+1}\leftarrow$ execute $a_t$ in the environment\;
  \lIf{the task is completed}{\textbf{break}}
  \Comment{task-progress tracking}
  $D_{t+1}\leftarrow(a_t,o_{t+1})$\;
  $Z_{t+1}\leftarrow f_\theta(Z_t,D_{t+1})$\;
  $B_{t+1}\leftarrow\mathcal E(u,H_{<t+1})$\;
  $H_{t+1}\leftarrow$ recent interactions up to step $t$\;
}
\Return{$\tau=(o_0,a_0,o_1,\ldots)$}
\end{algorithm}

\begin{algorithm}[!htb]
\caption{Training the trainable components}
\label{alg:training}
\small
\SetKwInOut{Input}{Require}\SetKwInOut{Output}{Ensure}\SetCommentSty{textrm}\SetKwComment{Comment}{$\triangleright$\ }{}
\Input{training trajectories $\{\tau_i\}$, skill library $\mathcal K$, frozen base model, frozen executor $\pi_{\eta^\star}$}
\Output{encoder $f_\theta$, navigator $g_\phi$, projection $P_\psi$}
\Comment{stage 1: history-compression pretraining}
\For{each epoch and each trajectory $\tau_i$}{
  $Z_{i,0}\leftarrow Z_{\mathrm{init}}$\;
  \For{each decision point $t$ of $\tau_i$}{
    $Z_{i,t+1}\leftarrow f_\theta(Z_{i,t},D_{i,t+1})$ \Comment*[r]{gradients truncated across steps}
    accumulate $\mathcal L_{\mathrm{comp}}$ (Eq.~\ref{eq:memory-training})\;
  }
  update $\theta$ with $\nabla\mathcal L_{\mathrm{comp}}$\;
}
save the pretrained encoder checkpoint\;
\Comment{stage 2: state-grounded navigation fine-tuning}
\For{each epoch and each trajectory $\tau_i$}{
  $Z_{i,0}\leftarrow Z_{\mathrm{init}}$\;
  \For{each decision point $t$ of $\tau_i$}{
    $Z_{i,t+1}\leftarrow f_\theta(Z_{i,t},D_{i,t+1})$ \Comment*[r]{gradients truncated across steps}
    $y_{i,t}\leftarrow$ uniform mass over the skills supporting the next action\;
    accumulate $\mathcal L_{\mathrm{nav}}$ (Eq.~\ref{eq:nav-training})\;
  }
  update $\theta,\phi$ with $\nabla\mathcal L_{\mathrm{nav}}$\;
}
select the checkpoint on held-out training scenarios\;
\Comment{stage 3: executor interface}
\For{each decision point $(i,t)$}{
  compute $B_{i,t}$, $Z_{i,t}$, and $\mathcal K_{i,t}$ with the trained $f_\theta,g_\phi$\;
  update $\psi$ with $\nabla\mathcal L_{\mathrm{proj}}$ (Eq.~\ref{eq:executor-training}); $\eta^\star$ stays fixed\;
}
\end{algorithm}

\subsection{Model Selection and Training Configuration}

To select a checkpoint that generalizes across task scenarios, we hold out scenario-disjoint validation trajectories from the training split and use them only for checkpoint selection. Table~\ref{tab:method-hparams} lists the configuration.

\section{Preliminary Study Details}
\label{app:preliminary-details}

This section supplements Section~\ref{sec:preliminary} with the comparison setup, metric definitions, and complete results behind Figure~\ref{fig:preliminary-execution-drift}.

\begin{table*}[!t]
\centering
\caption{\textbf{Preliminary comparison of retrieval timing and query content on AppWorld.} The skill library, base retriever, executor, skill-injection budget $C_K$, and interaction limit are fixed. Hit@3 measures stage-specific skill coverage at online decision points; TGC is reported in percent, and D1/D2/D3 denote the AppWorld metadata difficulty levels.}
\label{tab:preliminary-online-retrieval}
\footnotesize
\setlength{\tabcolsep}{3pt}
\begin{tabular*}{\textwidth}{@{\extracolsep{\fill}}lccccc@{}}
\toprule
Method & Hit@3 $\uparrow$ & TGC$_{\mathrm{D1}}$ $\uparrow$ & TGC$_{\mathrm{D2}}$ $\uparrow$ & TGC$_{\mathrm{D3}}$ $\uparrow$ & Avg.\ query tokens $\downarrow$ \\
\midrule
Initial Retrieval & 0.115 & 85.4 & 69.4 & \textbf{57.7} & \textbf{81} \\
Online Retrieval & \textbf{0.385} & \textbf{88.3} & \textbf{75.0} & 55.6 & 328 \\
Online Retrieval + Full History & 0.346 & 87.7 & 65.3 & 56.1 & 2{,}924 \\
\bottomrule
\end{tabular*}
\end{table*}

\subsection{Comparison Setup}
\label{app:preliminary-setup}

The preliminary comparison holds the skill library, base retriever, executor, skill-injection budget $C_K$, and interaction limit fixed. It varies when skills are acquired and which execution evidence is included in the retrieval query. The task-level results are reported on AppWorld, following the shared evaluation setting in Section~\ref{sec:experiments}.

\paragraph{Retrieval conditions.}
\textbf{Initial Retrieval} retrieves skills from the initial task instruction before execution and retains that initial selection. \textbf{Online Retrieval} updates its retrieved skill set at every decision point using the task instruction together with the local execution context: the preceding thought and the most recent $1{,}200$ characters of the current observation. \textbf{Online Retrieval + Full History} additionally supplies the complete preceding interaction history to the online query. \textbf{Online Retrieval + Structured History} instead prepends a short record of the task goal, the interfaces already called, and the failures observed so far. The structured-history condition tests the organization of textual history; it is not the full explicit--latent state mechanism of \textsc{JitSkill}.

\paragraph{Difficulty groups.}
D1, D2, and D3 correspond to values $1$, $2$, and $3$ of AppWorld's task-metadata difficulty field. They are benchmark-defined difficulty groups, not bins obtained from the number of execution steps or observed trajectory lengths.

\paragraph{Configuration.}
All conditions share the AppWorld skill library built from the official training split, the same base retriever, which fuses BM25 with all-MiniLM-L6-v2 dense scores by reciprocal rank fusion, and the same frozen executor, Qwen3.8-27B. Each decision point receives at most $k=3$ skills within a skill-injection budget of $C_K=1{,}200$ tokens, and an episode runs for at most $50$ interaction steps. The full-history query is capped at $C_Q=4{,}096$ tokens, keeping the most recent part when the history is longer. Query lengths are counted in the retriever's WordPiece tokens.

\paragraph{Retrieval-side and task-level measurements.}
Coverage, query cost, and the trace-level rates are measured by replaying every condition's retrieval on the same execution trajectories of the official test split, so all conditions are compared at identical decision points. Task goal completion is measured end-to-end by deploying each condition as the skill provider of the executor and running the official test split three times.

\subsection{Metrics and Annotation Scope}
\label{app:preliminary-metrics}

\paragraph{Stage-specific skill coverage (Hit@3).}
Hit@3 targets skill needs that emerge during execution. An evaluation event is the first step at which a skill becomes needed whose application the task request does not name, so no query formed before execution can be expected to anticipate it. Let $\mathcal{E}$ denote these events, $G_e$ the set of skills compatible with the need at event $e$, and $K^{(3)}_e$ the top-three skills supplied by the condition being evaluated. Then
\begin{equation}
\mathrm{Hit@3}=\frac{1}{|\mathcal{E}|}
\sum_{e\in\mathcal{E}}\mathbf{1}\!\left[K^{(3)}_e\cap G_e\neq\varnothing\right].
\end{equation}
For Initial Retrieval, coverage is assessed using the initial selection rather than a new query at that event. Hit@3 is an event-level coverage measure: it is not the fraction of correct skills among the top three, nor the probability of eventual task success.

\paragraph{Need annotation.}
The skills needed at a step are derived deterministically from the trajectory: a skill is needed when the interfaces it covers include an interface the step actually calls, excluding ubiquitous calls for documentation lookup, account-password lookup, and task completion. Each need is counted once, at the step where it first appears, so a need that spans several steps does not contribute repeated events. Because the labels follow from executed calls, the annotation requires no human judgment and is identical across conditions.

\paragraph{Task outcomes and query cost.}
Task goal completion (TGC) reports task-level completion, expressed as a percentage. TGC$_{\mathrm{D1}}$, TGC$_{\mathrm{D2}}$, and TGC$_{\mathrm{D3}}$ report this outcome within the three metadata difficulty groups. Average query tokens measure the length of the retrieval query, not the skill-injection payload or the executor's total token consumption.

\paragraph{Trace-level error rates.}
The diagnosis examines the top-ranked retrieved skill at every decision event that has at least one needed skill. Redundant exploration marks an event whose top-ranked skill was already retrieved earlier in the episode by the same condition and whose covered outcome has already been produced, that is, at least one of its interfaces has already been called; retrieving the same skill again is not sufficient on its own. Execution-consistency deviation marks an event whose top-ranked skill is needed neither at the current step nor at any later step of the episode, so it cannot advance any remaining requirement. These are event-level error rates, not task success rates conditioned on having retrieved a correct skill. All conditions are evaluated on the same $954$ decision events, derived with the same need annotation.

\subsection{Full Preliminary Results}
\label{app:preliminary-results}

\paragraph{Coverage, task completion, and query cost.}
Table~\ref{tab:preliminary-online-retrieval} provides the full comparison underlying Figure~\ref{fig:preliminary-execution-drift}b, including the full-history condition and query cost. Online Retrieval raises Hit@3 from $0.115$ to $0.385$ relative to Initial Retrieval. Its TGC improves by $2.9$ and $5.6$ percentage points on D1 and D2, but decreases by $2.1$ points on D3. Adding full history yields $0.346$ Hit@3 at an average query length of $2{,}924$ tokens, compared with $328$ tokens for Online Retrieval.

\begin{figure}[!htbp]
\centering
\includegraphics[width=\columnwidth]{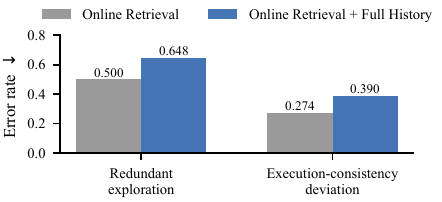}
\caption{\textbf{Trace-level diagnosis of online retrieval.} Error rates of the top-ranked retrieved skill over the same $954$ annotated decision events. Repeated retrieval counts as redundant only when earlier retrieval has already produced the same task-relevant outcome.}
\label{fig:preliminary-error-analysis}
\end{figure}

\paragraph{Error diagnosis.}
Figure~\ref{fig:preliminary-error-analysis} reports the aggregate-event diagnosis over $954$ decision events, a different scope from the matched long-horizon comparison in Figure~\ref{fig:preliminary-execution-drift}c. Adding full history increases redundant exploration by $14.8$ percentage points and execution-consistency deviation by $11.6$ points.

\paragraph{Structured-history comparison.}
Online Retrieval + Structured History raises Hit@3 to $0.415$ with $423$ query tokens, compared with $0.385$ and $328$ tokens for Online Retrieval and $0.346$ and $2{,}924$ tokens for Online Retrieval + Full History. However, its execution-consistency deviation rises to $45.7\%$, exceeding both Online Retrieval ($27.4\%$) and Online Retrieval + Full History ($39.0\%$). This aggregate-event result differs from the matched long-horizon comparison below; conclusions about the two scopes should be distinguished.

\paragraph{Matched long-horizon diagnosis.}
Figure~\ref{fig:preliminary-execution-drift}c reports a matched long-horizon comparison across the three online query conditions. Redundant exploration is $55.0\%$, $73.3\%$, and $51.6\%$ for Online Retrieval, Full History, and Structured History, respectively; execution-consistency deviation is $33.3\%$, $44.0\%$, and $30.2\%$. Thus, full history increases the two rates by $18.3$ and $10.7$ percentage points relative to local-context retrieval, whereas structured history reduces them by $3.4$ and $3.1$ points. Within this comparison, organizing history reduces observed errors, although substantial errors remain. These rates should not be attributed to the separate $954$-event aggregate diagnosis.

\FloatBarrier

\section{Experimental Setup Details}
\label{app:experiment-details}

\subsection{Baselines}
\label{app:baselines}

Each external method is re-implemented under the shared protocol of Section~\ref{sec:experiments}: the same skill library, retrieval backbone, executor, injection location, and per-step skill budget. Published end-to-end scores from web or desktop environments are therefore not transferred; the comparison isolates the decision rule that determines when and which skills reach the executor. Table~\ref{tab:baseline-design} groups the baselines by this decision rule: pre-execution retrieval, retrieval from the current local signal, and retrieval conditioned on a maintained execution state.

\subsubsection{CUA-Skill}
\label{app:baseline-cua-skill}

CUA-Skill~\citep{chen2026cuaskill} is a computer-use skill framework that represents reusable desktop operations as parameterized execution graphs. During execution, its planner generates retrieval queries from the user instruction, the current user-interface state, and an execution-memory buffer; it retrieves and re-ranks candidate skills, instantiates the arguments of the selected skill, and executes the resulting operation. Its memory buffer is initialized from the task instruction and then appends a concise summary of each executed skill and its observed outcome. The buffer therefore exposes prior successes, failures, and their contexts to later retrieval, re-ranking, and argument-instantiation decisions.

CUA-Skill addresses a complementary but different problem from \textsc{JitSkill}. Its parameterization and execution graphs aim to make a selected computer-use operation executable in the current interface. For example, argument instantiation fills operation-level slots such as a target UI element or coordinate, input text, file path, or application-specific option; the memory buffer helps the planner avoid repeating unsuccessful operations and recover from execution failures. These mechanisms improve how an already selected skill is instantiated and executed under local interface conditions.

In contrast, \textsc{JitSkill} addresses whether a skill should remain part of the task-level decision process as execution progresses. It maintains task-progress state that distinguishes unresolved requirements, satisfied conditions, reusable resources, failed routes, and active commitments. This state determines whether external support should be newly retrieved, whether previously available support should be reused or skipped, and whether a locally executable skill is compatible with the remaining task constraints. Thus, successful operation execution alone is insufficient: a skill can be correctly configured and executed while being redundant after a requirement has been satisfied, or inconsistent with a commitment established earlier in the trajectory. We include CUA-Skill as a strong execution-time retrieval baseline to distinguish operation-level execution support from task-progress-grounded skill lifecycle control.

\subsubsection{Reproduced Baselines}
\label{app:baseline-impl}

All reproduced baselines run under the shared protocol: the same frozen executor, the same skill library, the same injection location, $k=3$ skills per step, and the same skill-injection budget $C_K$. Each follows the default configuration of its paper, and any extra model call a method requires is issued to the same executor. To keep the retrieval backbone identical across methods, every method embeds skills with all-MiniLM-L6-v2.

\paragraph{SGDR.} At each step, the executor writes a short summary of the current state. Each skill is scored by a weighted combination of its similarity to the task goal and to this state summary (weight $0.5$), and the shortlisted skills are selected by maximal marginal relevance ($\lambda=0.7$).

\paragraph{SkillTrace.} Once per task, the instruction is decomposed into atomic sub-queries. Sub-queries are matched to skills by optimal bipartite assignment, and supporting skills are added by reverse personalized PageRank over the skill input--output dependency graph (restart probability $0.2$, $50$ iterations, $8$ linking candidates, linking threshold $0.6$). Matched skills are injected first, followed by supporting skills in PageRank order.

\paragraph{WebXSkill.} We follow its guided mode. Candidate skills are retrieved by embedding similarity in place of its URL-pattern graph, which has no counterpart outside web navigation, and the selected skill's step-by-step guidance is injected at the following decision points.

\paragraph{SkillZip.} Each skill is segmented by the executor into sections, each labeled with one of the nine execution roles. At the start of a task, sections and skills are seeded by reciprocal rank fusion against task anchors, missing core roles (intent, operation, verifier) are filled in, and the selection is rendered at the finest level of detail that fits the budget. Motif mining is omitted because the $37$-skill ALFWorld library contains no repeated cross-skill motifs.

\paragraph{FlowState.} At each step, the executor emits a state delta (add, update, or remove) over typed state nodes (preferences, knowledge, attributes, judgments, and artifacts) following its protocol; validated deltas are committed to the active state, whose summary serves as the retrieval query. The injected block contains the active state and the retrieved skills.

\paragraph{StructAgent.} The task is first decomposed into requirements, all marked pending. At each step, a verifier call reads the latest observation and marks requirements as verified or invalidated; only the verifier changes requirement status. The first pending requirement serves as the current subgoal for skill retrieval, and the block contains the requirement table, the subgoal, and the retrieved skills. Textual observations replace its GUI probes.

\paragraph{SoftSkill.} Each ALFWorld task type receives $32$ trainable virtual tokens, initialized from the embeddings of the corresponding skill description and trained with next-token prediction on successful trajectories from the official training split, with the executor frozen. Trajectories from the shared executor replace the original teacher model. The checkpoint is selected by success rate on held-out training games, and the virtual tokens are attached to the task turn through the serving engine's embedding interface.

\paragraph{CodeSkill.} The shared executor converts the skill library into CodeSkill's skill format and produces add, merge, and drop maintenance decisions over training-split trajectories; a Qwen3-8B skill manager is fine-tuned on these decisions with LoRA (rank $16$, learning rate $10^{-4}$, two epochs), rebalancing non-empty decisions so that the manager does not default to no-op. At evaluation time, the manager maintains the library online after each game, and skills are retrieved through task-level and event-level queries. We use the supervised stage of CodeSkill.

All reproduced methods share the same per-step skill budget $C_K$, so every injected block has the same size limit.

\paragraph{SKILL.state.} The append-only history is replaced by the triple of a procedural specification, a structured execution state, and the latest observation. The executor returns a state patch together with its action; the patch is validated against the state schema (up to three attempts, with the rejection returned to the executor) and merged into the state. The procedural specification is retrieved once from the task instruction.

\subsection{Implementation Details}
\label{app:implementation}

\paragraph{Direct-agent comparison.} Table~\ref{tab:direct-agent-details} reports all four no-skill agents considered during model selection. The main table retains the two Qwen models used by our method and executor protocol; the two architecture-matched alternatives are reported here for completeness. Each SkillsBench and $\tau^2$-Bench entry is the mean of three runs. All entries are success rates (\%).

\begin{table*}[t]
\centering
\caption{\textbf{Success rates (\%) of the no-skill direct agents.} SkillsBench contains $6$ Core, $53$ Extended, and $28$ Extreme tasks; ALFWorld contains $140$ in-distribution games; $\tau^2$-Bench contains $20$ airline, $40$ retail, and $40$ telecom tasks.}
\label{tab:direct-agent-details}
\renewcommand{\arraystretch}{1.08}
\footnotesize
\setlength{\tabcolsep}{1.5pt}
\begin{tabular*}{\textwidth}{@{\extracolsep{\fill}}l*{15}{c}@{}}
\toprule
\multirow{2}{*}{Model} & \multicolumn{4}{c}{SkillsBench} & \multicolumn{7}{c}{ALFWorld} & \multicolumn{4}{c}{$\tau^2$-Bench} \\
\cmidrule(lr){2-5}\cmidrule(lr){6-12}\cmidrule(lr){13-16}
& Core & Ext. & Extr. & SR & Pick & Look & Clean & Heat & Cool & Pick2 & SR & Airline & Retail & Telecom & Avg. \\
\midrule
Qwen3-8B & 0.00 & 0.00 & 3.57 & 1.15 & 71.43 & 30.77 & 29.63 & 0.00 & 0.00 & 8.33 & 27.86 & 15.00 & 2.50 & 1.67 & 6.39 \\
Llama-3.1-8B-Instruct & 0.00 & 0.00 & 0.00 & 0.00 & 11.43 & 15.38 & 0.00 & 6.25 & 0.00 & 8.33 & 6.43 & 11.67 & 3.33 & 7.50 & 7.50 \\
Qwen3.8-27B & 0.00 & 8.18 & 3.57 & 6.13 & 88.57 & 84.62 & 85.19 & 6.25 & 52.00 & 75.00 & 69.29 & 38.33 & 56.67 & 20.83 & 38.61 \\
Gemma-4-31B-IT & 16.67 & 6.29 & 3.57 & 6.13 & 95.24 & 66.67 & 65.43 & 0.00 & 50.67 & 77.78 & 65.00 & 41.62 & 64.17 & 22.50 & 42.76 \\
\bottomrule
\end{tabular*}
\end{table*}

\paragraph{Executor.} All skill-augmented methods share Qwen3.8-27B as a frozen executor, served with vLLM using FP8 weights and a $32{,}768$-token context. The same-scale direct agents are served with the same configuration and each model's official chat template and reasoning parser. On SkillsBench, the executor acts through native tool calls parsed by the model's official tool-call parser.

\paragraph{Skill injection.} Every method delivers its selected skills at the same prompt location under the same per-step budget. We use $k=3$ skills and $C_K=1{,}200$ tokens by default; Appendix~\ref{app:sensitivity} shows that \textsc{JitSkill} is robust to both settings.

\paragraph{Navigator.} The navigator is trained on trajectories from official training splits only. Its backbone is Qwen3-8B with LoRA adapters (rank $32$, $\alpha=64$, dropout $0.05$) on the attention projections and $16$ latent slots. At each step, the latent state is updated recurrently from the previous state, the executor's latest thought, and the most recent observation. The weight that combines the learned skill score with the relevance of the current step is selected by success rate on $200$ games of the official ALFWorld training split, and evaluation games are never used for selection.

\paragraph{Explicit record.} $B_t$ is extracted from the interaction history by deterministic rules. On AppWorld, it keeps the task goal, the six most recent API calls, and the six most recent failed calls. On ALFWorld and SkillsBench, it keeps the task goal, the completed actions (environment commands or tool calls) with a short excerpt of their observations, and the actions that failed. The record is packed into a fixed token budget.

\paragraph{Harnesses.} AppWorld runs use the official evaluation harness and the official task-goal and scenario-goal completion metrics. ALFWorld runs use the public text-world harness with the official in-distribution evaluation games, scoring a game as solved only when the environment reports task completion. SkillsBench runs use the official benchmark harness, in which each task executes in its own container and is graded by the task's own verifier.

\paragraph{Hardware.} All executors and navigators run on NVIDIA GPUs with at least $48$\,GB of memory. All executor instances use identical serving parameters.

\paragraph{Benchmarks.}
\begin{itemize}
\item \textbf{AppWorld}: official \texttt{test\_normal} split ($168$ tasks) and \texttt{test\_challenge} split ($417$ tasks), at most $50$ interaction steps; the skill library is built from the official training split.
\item \textbf{ALFWorld}: the $140$ in-distribution evaluation games, at most $30$ steps, with the $37$ ALFWorld skills from the public skill collection used by prior work; results are reported by the six task types.
\item \textbf{SkillsBench}: all $87$ public tasks under the official harness and each task's native time budget, with the skills bundled with the benchmark; difficulty groups follow each task's own label.
\item \textbf{$\tau^2$-Bench}: the official test split of the airline ($20$), retail ($40$), and telecom ($40$) domains, with the same user simulator, executor, step limit ($40$), and seed as the direct-agent rows, three runs. Since the benchmark ships no skill library, we follow prior skill-based evaluations on $\tau^2$-Bench and distill $10$--$12$ skills per domain from the executor's successful trajectories on the official training split, together with each domain's policy document and tool schema; test tasks are never used. Skills are appended to the system prompt, and the fusion weight is selected on the training split.
\end{itemize}

\paragraph{Runs and statistics.} Each AppWorld and ALFWorld configuration is run three times with fixed seeds; each SkillsBench task runs in its own container for up to several hours and is run three times. Paired comparisons use an exact sign test over tasks and a paired bootstrap over tasks.

\section{Additional Results}
\label{app:additional-results}

\subsection{Dynamic Execution-State Maintenance on ALFWorld}
\label{app:dynamic-state-analysis}

Dynamic execution-state maintenance is a natural strategy for long-horizon agents: it records what has happened so that later decisions need not reconstruct task progress from the entire interaction history. The methods differ, however, in what their state represents and in whether it conditions subsequent skill retrieval and execution. Table~\ref{tab:dynamic-state-alfworld} compares the dynamic state-maintenance methods under the common ALFWorld protocol.

\paragraph{State representations provide different decision signals.} StructAgent maintains verifiable requirement status, and FlowState preserves state--evidence pairs for later reuse. \textsc{SKILL.state} replaces the history with a procedural specification, a structured state, and the latest observation, but retrieves its procedural specification only once from the initial task instruction. Their overall success rates range from $52.38\%$ to $86.19\%$.

\paragraph{Progress-grounded retrieval and execution are complementary.} \textsc{JitSkill} reaches $92.38\%$ by maintaining both an explicit progress record and a latent progress representation, then using the shared state to condition both which skills are retrieved and how the executor applies them. This comparison indicates that retaining execution state alone is insufficient when skill decisions must remain consistent with completed work and unresolved requirements.

\begin{table*}[!t]
\centering
\caption{\textbf{Success rates (\%) of dynamic execution-state methods on ALFWorld.} Bold and underline mark the best and second-best values.}
\label{tab:dynamic-state-alfworld}
\renewcommand{\arraystretch}{1.05}
\footnotesize
\setlength{\tabcolsep}{3pt}
\begin{tabular*}{\textwidth}{@{\extracolsep{\fill}}lccccccc@{}}
\toprule
Method & Pick & Look & Clean & Heat & Cool & Pick2 & SR \\
\midrule
StructAgent & \underline{95.24} & 53.85 & 80.25 & \underline{81.25} & 52.00 & 87.50 & 77.86 \\
FlowState & \textbf{98.10} & \underline{89.74} & \underline{83.95} & 72.92 & \underline{69.33} & \textbf{95.83} & \underline{86.19} \\
\textsc{SKILL.state} & 64.76 & 41.03 & 62.96 & 52.08 & 42.67 & 38.89 & 52.38 \\
\midrule
\textsc{JitSkill} & \textbf{98.10} & \textbf{94.87} & \textbf{90.12} & \textbf{83.33} & \textbf{89.33} & \underline{94.44} & \textbf{92.38} \\
\bottomrule
\end{tabular*}
\end{table*}

\subsection{Performance by Task Difficulty}

To examine where state-grounded skill retrieval matters most, we split the AppWorld frozen-executor results by the benchmark's own difficulty field into difficulty levels $1$, $2$, and $3$ ($57$ / $48$ / $63$ tasks). The labels are read from task metadata, so no method can relabel its own tasks. Table~\ref{tab:horizon} reports task goal completion in each group.

\paragraph{The advantage of \textsc{JitSkill} grows with task difficulty.} \textsc{JitSkill} is the strongest method in every group, and its margin over step-wise retrieval widens from $+1.2$ at difficulty level $1$ to $+5.6$ at difficulty level $2$ and $+9.0$ at difficulty level $3$. This pattern indicates a larger benefit from tracked task progress in the higher-difficulty group.

\paragraph{Local retrieval degrades at difficulty level $3$.} Step-wise retrieval helps at difficulty levels $1$ and $2$ but falls below the no-skill agent at difficulty level $3$ ($55.6$ vs.\ $61.9$), and removing the task instruction from its query drops it further to $48.7$. These results indicate that local retrieval is insufficient to improve completion in the highest-difficulty group, whereas \textsc{JitSkill} remains above the no-skill agent in every group.

\paragraph{Appending more history is not a substitute for tracked state.} Adding the raw textual history or the full history to the retrieval query does not recover the loss at difficulty level $3$ ($57.7$ and $56.1$), and the full history even reduces completion at difficulty level $2$ to $65.3$. These results show that adding historical context alone does not resolve this performance gap; \textsc{JitSkill} instead conditions retrieval on a compact state of task progress and reaches $64.6$ at difficulty level $3$.

\begin{table}[!htbp]
\centering
\caption{\textbf{Task goal completion (\%) by difficulty level on AppWorld.} Difficulty levels follow the task metadata of the official test split; three runs with a frozen executor.}
\label{tab:horizon}
\footnotesize
\setlength{\tabcolsep}{3pt}
\begin{tabular*}{\columnwidth}{@{\extracolsep{\fill}}lccc@{}}
\toprule
Method & TGC$_{\mathrm{D1}}$ $\uparrow$ & TGC$_{\mathrm{D2}}$ $\uparrow$ & TGC$_{\mathrm{D3}}$ $\uparrow$ \\
\midrule
No skills & 83.0 & 70.1 & 61.9 \\
Pre-execution retrieval & 85.4 & 69.4 & 57.7 \\
Step-wise retrieval & 88.3 & 75.0 & 55.6 \\
\quad without instruction & 87.7 & 69.4 & 48.7 \\
\quad + raw textual history & 86.5 & 68.8 & 57.7 \\
\quad + full history & 87.7 & 65.3 & 56.1 \\
\textsc{JitSkill} & \textbf{89.5} & \textbf{80.6} & \textbf{64.6} \\
\bottomrule
\end{tabular*}
\end{table}

\subsection{Sensitivity to the Number of Skills and the Injection Budget}
\label{app:sensitivity}
\label{app:param-sensitivity}

\begin{figure}[t]
\centering
\includegraphics[width=\columnwidth]{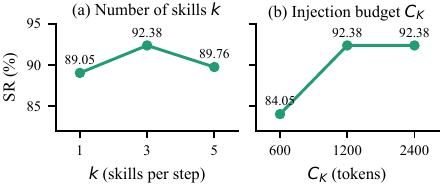}
\caption{Sensitivity of \textsc{JitSkill} to (a) the number of injected skills $k$ ($C_K=1{,}200$) and (b) the injection budget $C_K$ ($k=3$) on ALFWorld, averaged over three runs.}
\label{fig:app-sensitivity}
\end{figure}

We study the two hyper-parameters that control how many retrieved skills reach the executor: the number of injected skills $k$ and the injection budget $C_K$. We vary $k$ at the default budget and $C_K$ at the default $k$ on ALFWorld, keeping all other settings fixed (Figure~\ref{fig:app-sensitivity}): $k=1,3,5$ reach $89.05\%$, $92.38\%$, and $89.76\%$, and $C_K=600$, $1{,}200$, and $2{,}400$ reach $84.05\%$, $92.38\%$, and $92.38\%$.

\paragraph{\textsc{JitSkill} is robust to the number of injected skills.} The default $k=3$ performs best ($92.38\%$), and every setting from $k=1$ to $k=5$ stays above $89\%$. Because the navigator ranks skills against the tracked task state, the most relevant skill is already at the top of the list: with a single skill, \textsc{JitSkill} still reaches $89.05\%$.

\paragraph{A moderate budget suffices.} Halving the budget to $C_K=600$ lowers success to $84.05\%$, because the progress record and one skill document no longer fit together and the procedure is cut short. Doubling it to $C_K=2{,}400$ brings no further gain ($92.38\%$ in both cases): once the skills that match the current stage are delivered in full, additional procedural text adds nothing. The navigator therefore makes efficient use of a moderate budget, and we use $C_K=1{,}200$ in all comparisons so that every method operates under the same budget.

\subsection{Efficiency on ALFWorld}
\label{app:efficiency-alfworld}

We complement the AppWorld cost analysis in the main text with the interaction length on ALFWorld and the overhead of the navigator itself. Table~\ref{tab:alfw-efficiency} reports the average number of environment steps per game.

\paragraph{\textsc{JitSkill} succeeds more while acting less.} \textsc{JitSkill} takes $24.0\%$ fewer steps than the no-skill agent and $11.2\%$ fewer than SGDR, while achieving the highest success rate. Its gains therefore do not come from longer interaction: skills that match the current stage of the task let the executor reach the goal along a shorter path. In contrast, SKILL.state, which rewrites a structured state at every step, takes the most steps.

\paragraph{The navigator adds little overhead.} With the navigator on the executor's GPU, building the skill block takes about $0.13$\,s per step on ALFWorld and about $1.1$\,s per step on AppWorld, a small fraction of an executor call. The latent state compresses a median interaction history of about $4{,}600$ tokens into $16$ slots, and training the navigator takes about $11$ minutes on a single GPU.

\begin{table}[!htbp]
\centering
\caption{\textbf{Interaction length on ALFWorld.} Average environment steps per game on the $140$ in-distribution games (three runs, at most $30$ steps).}
\label{tab:alfw-efficiency}
\footnotesize
\setlength{\tabcolsep}{3pt}
\begin{tabular*}{\columnwidth}{@{\extracolsep{\fill}}lcc@{}}
\toprule
Method & Steps / game $\downarrow$ & SR (\%) $\uparrow$ \\
\midrule
No skills & 17.57 & 70.24 \\
SGDR & 15.03 & 87.14 \\
SkillTrace & 15.53 & 81.19 \\
SKILL.state & 21.10 & 52.38 \\
\textsc{JitSkill} & \textbf{13.35} & \textbf{92.38} \\
\bottomrule
\end{tabular*}
\end{table}

\begin{figure}[!htbp]
\centering
\includegraphics[width=\columnwidth]{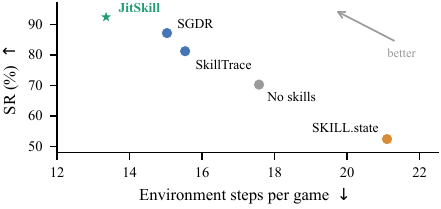}
\caption{\textbf{Success rate versus interaction length on ALFWorld.} Results on the $140$ in-distribution games (three runs); the upper left is better.}
\label{fig:efficiency-scatter-appendix}
\end{figure}

\paragraph{Execution cost under a fixed budget.} Table~\ref{tab:ts-cost} keeps the skill-injection budget fixed, so the reported payload differences arise from episode length and retrieval policy rather than an unrestricted context advantage. The whole-library row injects every skill and serves only as a reference for the payload size: it saturates the context window, so $21.2\%$ of its model calls return an error, and it is not a comparable baseline.

\begin{table*}[!t]
\centering
\caption{\textbf{Execution cost per episode on AppWorld.} Results on the official test split (three runs). The whole-library row is a payload reference rather than a comparable baseline.}
\label{tab:ts-cost}
\footnotesize
\setlength{\tabcolsep}{3pt}
\begin{tabular*}{\textwidth}{@{\extracolsep{\fill}}lccc@{}}
\toprule
Method & Steps / ep. & \makecell[c]{Skill tokens /\\step} & \makecell[c]{Skill tokens /\\ep.} \\
\midrule
No skills & 17.8 & 0 & 0 \\
Whole library & 4.4 & 27{,}828 & 122{,}465 \\
Pre-execution retrieval & 15.5 & 1{,}072 & 16{,}605 \\
Step-wise retrieval & 14.6 & 1{,}107 & 16{,}193 \\
\quad + raw textual history & 14.7 & 1{,}212 & 17{,}765 \\
\textsc{JitSkill} & 15.8 & 1{,}209 & 19{,}051 \\
\bottomrule
\end{tabular*}
\end{table*}

\subsection{Training Objectives and Curves}
\label{app:training-curves}

\textsc{JitSkill} trains task-state tracking and navigation in two stages. It first pretrains the encoder with the history-compression objective $\mathcal L_{\mathrm{comp}}$, then jointly fine-tunes the encoder and navigator with $\mathcal L_{\mathrm{nav}}$. The executor-side projection is trained subsequently with $\mathcal L_{\mathrm{proj}}$ while the executor remains frozen. We retrain the history compressor with each term of $\mathcal L_{\mathrm{comp}}$ removed or reweighted, keeping data, steps, and seed fixed, and evaluate on held-out scenarios (Table~\ref{tab:lmem-weights}).

\paragraph{Each term of the compression objective is necessary.} Each term protects its own function. Without $\mathcal L_{\mathrm{align}}$, the latent state drifts away from the reference distribution (JSD rises from $0.085$ to $0.500$); without $\mathcal L_{\mathrm{disc}}$, the latent state can no longer identify its own episode and retrieval falls to chance; and halving $\lambda_{\mathrm{rec}}$ degrades reconstruction. The three terms are thus complementary rather than redundant.

\paragraph{The default weights lie on a flat optimum.} Under the full objective, only doubling $\lambda_{\mathrm{rec}}$ matches the default weights, and every other setting is worse. The method therefore does not depend on a finely tuned weighting.

\begin{table*}[!t]
\centering
\caption{\textbf{Effect of each term of $\mathcal L_{\mathrm{comp}}$ on held-out scenarios.} Each row reports the quantity that the changed term controls, under the default weights and under the variant.}
\label{tab:lmem-weights}
\footnotesize
\setlength{\tabcolsep}{3pt}
\begin{tabular*}{\textwidth}{@{\extracolsep{\fill}}llcc@{}}
\toprule
Variant & Affected quantity & Default & Variant \\
\midrule
w/o $\mathcal L_{\mathrm{align}}$ & alignment JSD $\downarrow$ & \textbf{0.085} & 0.500 \\
w/o $\mathcal L_{\mathrm{disc}}$ & episode retrieval top-1 $\uparrow$ (chance $0.085$) & \textbf{0.337} & 0.072 \\
$\lambda_{\mathrm{rec}}\times 0.5$ & reconstruction NLL $\downarrow$ & \textbf{0.537} & 0.588 \\
\midrule
\multicolumn{2}{l}{Full objective $\mathcal L_{\mathrm{comp}}$ $\downarrow$ (default weights)} & 0.523 & 0.518--0.620 \\
\bottomrule
\end{tabular*}
\end{table*}

\begin{figure}[!htbp]
\centering
\includegraphics[width=\linewidth]{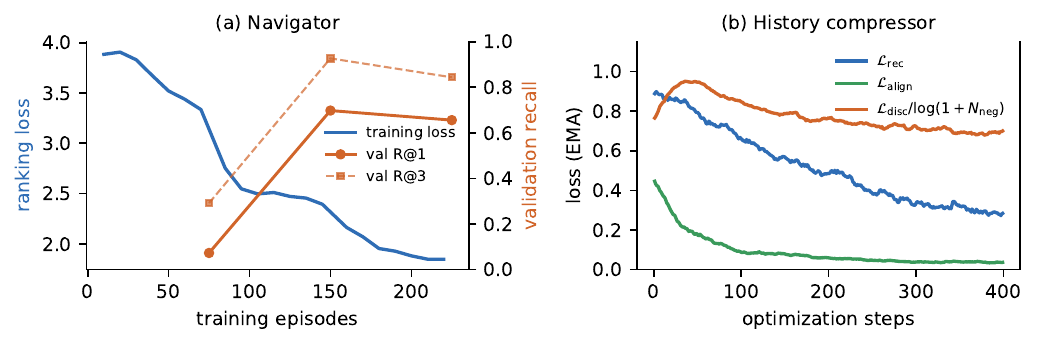}
\caption{\textbf{Training curves.} (a) Navigator ranking loss and validation recall of the next-step skill. (b) Terms of the history-compression objective under the default weights (exponential moving average).}
\label{fig:training-curves}
\end{figure}

\paragraph{Both state-training stages train stably and quickly.} Figure~\ref{fig:training-curves} shows the encoder pretraining and the subsequent navigation fine-tuning. The navigator's ranking loss decreases steadily, and validation recall of the next-step skill rises from $0.07$ to $0.70$ R@1 and $0.93$ R@3 within two epochs; this checkpoint is used in all experiments. All three terms of the history-compression objective decrease smoothly; the discriminative term is normalized by its chance level $\log(1+N_{\mathrm{neg}})$ because the negative queue fills during the first steps.

\subsection{Generalization to a Different Executor}
\label{app:executor-transfer}

\textsc{JitSkill} separates task-state tracking and navigation from the frozen executor. When a target executor requires a different latent-state interface, we keep the pretrained encoder and navigator fixed and train only its lightweight projection on official training trajectories. We test generalization across held-out environments and a different executor without using evaluation data for training or selection.

\paragraph{No evaluation data is used for training or selection.} The navigator is trained on official training splits, and every hyper-parameter is selected on training data, so the results in Table~\ref{tab:ts-main} measure generalization to held-out evaluation tasks.

\begin{figure}[!htbp]
\centering
\includegraphics[width=\columnwidth]{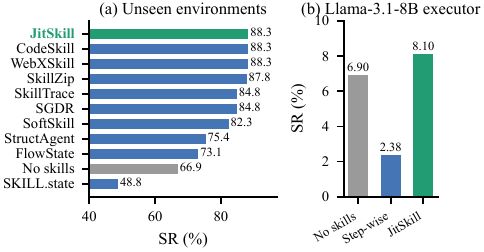}
\caption{\textbf{Generalization under held-out evaluation.} (a) Success rate on unseen ALFWorld environments ($134$ games) without re-tuning. (b) Success rate with Llama-3.1-8B-Instruct as the executor; the navigator remains unchanged and any executor-side interface is trained only on official training trajectories. Both average three runs.}
\label{fig:generalization}
\end{figure}

\paragraph{The navigator transfers to a different executor.} We further replace the executor with Llama-3.1-8B-Instruct while keeping the navigator unchanged, so the navigator now serves an executor it was never trained with. Each skill-augmented method selects its own hyper-parameter on $200$ games of the official ALFWorld training split under the same rule, and all methods share one system prompt that describes only the environment's action format. As shown in Figure~\ref{fig:generalization}b (three runs: $6.43$, $9.29$, and $8.57\%$ for \textsc{JitSkill}; $2.14$, $5.00$, and $0.00\%$ for step-wise retrieval; $7.86$, $3.57$, and $9.29\%$ for the no-skill agent), \textsc{JitSkill} achieves the highest success rate, whereas step-wise retrieval falls below the no-skill agent: for a smaller executor, skills that do not match the current stage of the task act as distractors, and state-grounded selection avoids them.

\subsection{Generalization to Unseen ALFWorld Environments}
\label{app:alfw-unseen}

We further evaluate on the official out-of-distribution split of ALFWorld (\texttt{valid\_unseen}, $134$ games), whose rooms do not appear in the in-distribution games. Every setting is identical to the main ALFWorld experiment: the same frozen executor, skill library, navigator, skill budget, prompt, and $30$-step limit; nothing is re-tuned on the unseen split. Table~\ref{tab:alfw-unseen} reports success rates averaged over three runs.

\paragraph{\textsc{JitSkill} remains among the strongest methods on unseen environments.} \textsc{JitSkill} reaches $88.31\%$, matching the highest overall success rate in Table~\ref{tab:alfw-unseen}, and improves on the no-skill agent by $21.4$ points. The other methods that maintain an execution state transfer far less well: StructAgent, FlowState, and SKILL.state, whose state is rewritten by the executor at every step, reach only $75.37\%$, $73.13\%$, and $48.76\%$. Because the navigator ranks skills against the tracked task progress rather than against surface features of a particular room, the selected procedures remain appropriate when the environment layout changes.

\paragraph{The gains concentrate on multi-stage tasks.} The largest improvements over the no-skill agent appear on Heat ($2.9\%\rightarrow78.3\%$) and Cool ($54.0\%\rightarrow92.1\%$), which require the agent to find an object, carry it to an appliance, operate the appliance, and then place the object. These tasks span several stages whose skill needs differ, and conditioning retrieval on the progress made so far supplies the procedure for the stage that comes next.

\begin{table*}[!t]
\centering
\caption{\textbf{Success rate (\%) on unseen ALFWorld environments.} Official out-of-distribution split ($134$ games: Pick $24$, Look $18$, Clean $31$, Heat $23$, Cool $21$, Pick2 $17$), averaged over three runs; all settings follow the main experiment without re-tuning.}
\label{tab:alfw-unseen}
\footnotesize
\setlength{\tabcolsep}{3pt}
\begin{tabular*}{\textwidth}{@{\extracolsep{\fill}}lccccccc@{}}
\toprule
Method & Pick & Look & Clean & Heat & Cool & Pick2 & SR $\uparrow$ \\
\midrule
No skills (Qwen3.8-27B) & 91.7 & 85.2 & 81.7 & 2.9 & 54.0 & 88.2 & 66.92 \\
SGDR & 88.9 & 85.2 & 87.1 & 71.0 & 90.5 & 86.3 & 84.83 \\
WebXSkill & 95.8 & \textbf{96.3} & 90.3 & 73.9 & 90.5 & 82.4 & \textbf{88.31} \\
SkillTrace & 97.2 & 90.7 & 82.8 & 63.8 & 84.1 & 94.1 & 84.83 \\
SoftSkill & 91.7 & 88.9 & 90.3 & 72.5 & 74.6 & 70.6 & 82.34 \\
CodeSkill & 95.8 & \textbf{96.3} & 88.2 & \textbf{85.5} & 74.6 & 90.2 & \textbf{88.31} \\
SkillZip & 94.4 & 92.6 & \textbf{95.7} & 66.7 & 88.9 & 86.3 & 87.81 \\
StructAgent & \textbf{98.6} & 72.2 & 69.9 & 60.9 & 57.1 & \textbf{98.0} & 75.37 \\
FlowState & 79.2 & \textbf{96.3} & 71.0 & 47.8 & 66.7 & 86.3 & 73.13 \\
SKILL.state & 51.4 & 48.1 & 59.1 & 43.5 & 54.0 & 27.5 & 48.76 \\
\midrule
\textsc{JitSkill} & 88.9 & 92.6 & 89.2 & 78.3 & \textbf{92.1} & 90.2 & \textbf{88.31} \\
\bottomrule
\end{tabular*}
\end{table*}

\subsection{Trajectory-level Diagnostics}

Beyond task completion, we examine how execution unfolds on AppWorld with two trajectory-level measures. Action calls per solved task divides the average number of API calls per episode, excluding documentation lookups, by the fraction of tasks completed, so it measures how much execution effort each success costs. Completion after feedback is the fraction of episodes that still complete the task after encountering at least one error observation, which reflects whether execution returns to the main line once something goes wrong.

\paragraph{\textsc{JitSkill} spends the least effort per success.} \textsc{JitSkill} needs $27.4$ action calls per solved task, $13\%$ fewer than the no-skill agent and $16\%$ fewer than step-wise retrieval. Tracked progress keeps the executor from repeating operations it has already completed.

\paragraph{\textsc{JitSkill} recovers best after errors.} After an error observation, \textsc{JitSkill} still completes $72.8\%$ of episodes, whereas both retrieval baselines recover less often than the no-skill agent ($68.8\%$ and $64.2\%$), and appending raw textual history lowers recovery further ($62.7\%$). Because the explicit record retains failed actions, the navigator can steer the executor toward skills that address the failure instead of retrieving support for the action that just failed.

\begin{table}[!htbp]
\centering
\caption{\textbf{Trajectory-level diagnostics on AppWorld.} Results on the official test split (three runs) with a frozen executor.}
\footnotesize
\setlength{\tabcolsep}{3pt}
\begin{tabular*}{\columnwidth}{@{\extracolsep{\fill}}lcc@{}}
\toprule
Method & \makecell[c]{Action calls /\\solved task $\downarrow$} & \makecell[c]{Completion after\\feedback (\%) $\uparrow$} \\
\midrule
No skills & 31.6 & 71.6 \\
Pre-execution retrieval & 32.7 & 68.8 \\
Step-wise retrieval & 32.5 & 64.2 \\
\quad + raw textual history & 30.3 & 62.7 \\
\textsc{JitSkill} & \textbf{27.4} & \textbf{72.8} \\
\bottomrule
\end{tabular*}
\end{table}

\subsection{Benchmark and Skill-Library Statistics}
\label{app:bench-stats}

Table~\ref{tab:bench-stats} summarizes the evaluation sets and the skill libraries shared by all skill-augmented methods. Skill lengths are counted with the executor's tokenizer.

\paragraph{The three benchmarks impose different retrieval demands.} ALFWorld pairs a compact library of $37$ procedural skills with long-horizon embodied tasks, so the challenge is to retrieve the procedure that fits the current stage of a multi-step task. SkillsBench ships a much larger library ($195$ skills) whose documents are long (median $1{,}121$ tokens, up to $8{,}199$), so skills cannot be injected in full: each selected skill is delivered as a compact pointer to its full document, and selection among many candidates becomes the main difficulty. $\tau^2$-Bench uses short, domain-specific skills ($245$ tokens on average) inside multi-turn conversations with a simulated user, where the relevant procedure changes as the dialogue moves from verification to lookup to the requested modification.

\begin{table*}[!t]
\centering
\caption{\textbf{Evaluation sets and skill libraries.} Test episodes are broken down by split, difficulty tier, or domain. Skill lengths are counted in executor tokens.}
\label{tab:bench-stats}
\footnotesize
\setlength{\tabcolsep}{3pt}
\begin{tabular*}{\textwidth}{@{\extracolsep{\fill}}lclcccc@{}}
\toprule
Benchmark & \makecell[c]{Test\\episodes} & Breakdown & \makecell[c]{Step\\limit} & \#Skills & \makecell[c]{Skill length\\(mean / median)} & \makecell[c]{Skill length\\(max)} \\
\midrule
ALFWorld & $140$ / $134$ & seen / unseen environments & $30$ & $37$ & $572$ / $567$ & $1{,}010$ \\
SkillsBench & $87$ & Core $6$, Extended $53$, Extreme $28$ & task-specific & $195$ & $1{,}517$ / $1{,}121$ & $8{,}199$ \\
$\tau^2$-Bench & $100$ & Airline $20$, Retail $40$, Telecom $40$ & $40$ & $33$ ($12$ / $11$ / $10$) & $245$ / $248$ & $298$ \\
\bottomrule
\end{tabular*}
\end{table*}

\subsection{Run-to-Run Variability}
\label{app:run-variability}

Table~\ref{tab:run-variability} reports the success rate of \textsc{JitSkill} in each of the three runs with the shared frozen executor.

\paragraph{The improvements are stable across runs.} On in-distribution ALFWorld, the three runs differ by less than one point ($92.14$--$92.86\%$), and every run exceeds the strongest baseline average ($90.48\%$). On SkillsBench, every run solves at least $13$ of the $87$ tasks, so even the weakest run ($14.94\%$) is above the average of every baseline in Table~\ref{tab:ts-main}. On $\tau^2$-Bench, the run-to-run standard deviation ($2.50$) is small relative to the gain over the direct agent with the same executor ($+10.6$ points), so the improvement is not an artifact of a single favorable run.

\begin{table}[!htbp]
\centering
\caption{\textbf{Run-to-run variability of \textsc{JitSkill}.} Success rate (\%) of each run and the mean $\pm$ sample standard deviation.}
\label{tab:run-variability}
\footnotesize
\setlength{\tabcolsep}{3pt}
\begin{tabular*}{\columnwidth}{@{\extracolsep{\fill}}lcccc@{}}
\toprule
Benchmark & Run 1 & Run 2 & Run 3 & Mean $\pm$ std \\
\midrule
ALFWorld (seen) & 92.14 & 92.86 & 92.14 & $92.38 \pm 0.41$ \\
ALFWorld (unseen) & 90.30 & 88.81 & 85.82 & $88.31 \pm 2.28$ \\
SkillsBench & 14.94 & 14.94 & 17.24 & $15.71 \pm 1.33$ \\
$\tau^2$-Bench (avg.) & 46.67 & 51.67 & 49.17 & $49.17 \pm 2.50$ \\
\bottomrule
\end{tabular*}
\end{table}

\subsection{How the Skill Library Is Used}
\label{app:skill-usage}

We next examine which skills reach the executor during execution (Table~\ref{tab:skill-usage}). A switch is a step at which the top-ranked injected skill differs from the previous step.

\paragraph{\textsc{JitSkill} re-selects skills as the task progresses.} Pre-execution retrieval fixes its selection before the first action and never switches, and step-wise retrieval switches only $3.33$ times per ALFWorld episode, because consecutive observations look alike even when the task has moved to a new stage. \textsc{JitSkill} switches $7.21$ times per episode and uses $7.31$ distinct skills per episode, more than twice as many as step-wise retrieval. Every injected block stays within the $1{,}200$-token budget, so these switches do not enlarge the context; they replace the injected procedure when the tracked progress indicates that the next stage needs a different one.

\paragraph{The whole library stays in use.} \textsc{JitSkill} draws on $36$ of the $37$ ALFWorld skills, $178$ of the $195$ SkillsBench skills, and all $33$ $\tau^2$-Bench skills, whereas pre-execution retrieval touches only $18$ ALFWorld skills. On SkillsBench, the five most frequently selected skills each account for only $3$--$4.4\%$ of selections, so the navigator does not collapse onto a few generic documents. On $\tau^2$-Bench, the most frequent skills are exactly the procedures for the requested operations, such as modifying or cancelling a reservation and exchanging delivered items.

\begin{table*}[!t]
\centering
\caption{\textbf{Skill usage per episode.} Averages over three runs; distinct skills count every skill injected at any position.}
\label{tab:skill-usage}
\footnotesize
\setlength{\tabcolsep}{3pt}
\begin{tabular*}{\textwidth}{@{\extracolsep{\fill}}llccc@{}}
\toprule
Benchmark & Method & \makecell[c]{Skills used\\in library} & \makecell[c]{Switches /\\episode} & \makecell[c]{Distinct skills /\\episode} \\
\midrule
\multirow{3}{*}{ALFWorld (seen)} & Pre-execution retrieval & 18/37 & 0.00 & 1.73 \\
& Step-wise retrieval & 34/37 & 3.33 & 4.34 \\
& \textsc{JitSkill} & \textbf{36/37} & \textbf{7.21} & \textbf{7.31} \\
\midrule
SkillsBench & \textsc{JitSkill} & 178/195 & 11.27 & -- \\
$\tau^2$-Bench & \textsc{JitSkill} & 33/33 & 6.22 & 8.42 \\
\bottomrule
\end{tabular*}
\end{table*}

\subsection{Where the Gains Come From}
\label{app:gain-breakdown}

\paragraph{Gains grow with the number of task stages.} Table~\ref{tab:stage-gain} groups the in-distribution ALFWorld games by task type. On single-stage tasks (Pick and Look), where the agent only needs to find and place an object, \textsc{JitSkill} improves the no-skill agent by $9.7$ points. On multi-stage tasks (Clean, Heat, Cool, and Pick2), which require the agent to carry an object through an appliance or to repeat a sub-task, the improvement triples to $30.1$ points, led by Heat ($+77.1$) and Cool ($+37.3$). \textsc{JitSkill} improves every one of the six task types.

\begin{table}[!htbp]
\centering
\caption{\textbf{Success rate (\%) by task type on ALFWorld (seen).} No skills is the Qwen3.8-27B direct agent of Table~\ref{tab:ts-main}.}
\label{tab:stage-gain}
\footnotesize
\setlength{\tabcolsep}{3pt}
\begin{tabular*}{\columnwidth}{@{\extracolsep{\fill}}llccc@{}}
\toprule
Stages & Type & No skills & \textsc{JitSkill} & $\Delta$ \\
\midrule
\multirow{3}{*}{Single} & Pick & 88.57 & 98.10 & $+9.5$ \\
& Look & 84.62 & 94.87 & $+10.3$ \\
& Combined & 87.50 & 97.22 & $+9.7$ \\
\midrule
\multirow{5}{*}{Multiple} & Clean & 85.19 & 90.12 & $+4.9$ \\
& Heat & 6.25 & 83.33 & $+77.1$ \\
& Cool & 52.00 & 89.33 & $+37.3$ \\
& Pick2 & 75.00 & 94.44 & $+19.4$ \\
& Combined & 59.78 & 89.86 & $+30.1$ \\
\bottomrule
\end{tabular*}
\end{table}

\begin{figure}[!htbp]
\centering
\includegraphics[width=\columnwidth]{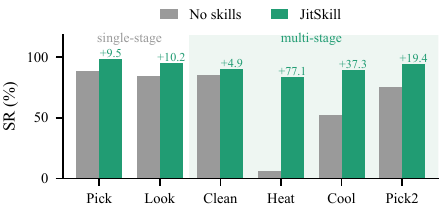}
\caption{\textbf{Gains by task type on ALFWorld.} Success rates of the no-skill agent and \textsc{JitSkill} with the same executor (three runs); numbers above the bars are the gains.}
\label{fig:stage-gain}
\end{figure}

\begin{figure}[!htbp]
\centering
\includegraphics[width=\columnwidth]{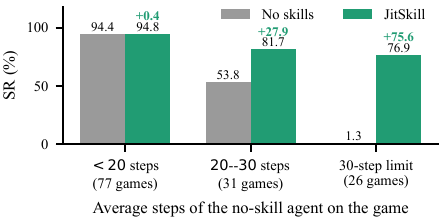}
\caption{\textbf{Success rate by task length on ALFWorld (unseen).} Games are grouped by the average number of steps the no-skill agent takes (three runs per game); green numbers are the gains of \textsc{JitSkill}.}
\label{fig:length-gain}
\end{figure}

\paragraph{Gains are largest on the games the executor cannot finish alone.} On the unseen split, we group games by how many steps the no-skill agent spends on them (Figure~\ref{fig:length-gain}). Games that the no-skill agent finishes in under $20$ steps are already almost always solved, and \textsc{JitSkill} keeps them solved. As the no-skill agent needs longer, the gap widens to $+28.0$ points, and on the $26$ games where it hits the $30$-step limit in all three runs, its success rate is $1.3\%$ while \textsc{JitSkill} succeeds in $76.9\%$ of attempts. Tracked progress thus matters most exactly where the executor otherwise loses its place in a long task.

\section{Case Study: Progress-Aware Skill Selection in an Unseen Environment}
\label{app:case-study}
\begingroup\sloppy

We trace one episode from the ALFWorld unseen split to show how the progress record changes which skills reach the executor. The game is a Heat task in a kitchen that does not appear in training: ``put a hot apple in fridge.'' \textsc{JitSkill} solves it in all three runs ($14$, $8$, and $12$ steps), whereas the no-skill executor fails all three runs and exhausts the $30$-step limit each time. All skill identifiers, progress records, actions, and observations below are copied from the execution logs.

\paragraph{Skill selection follows the stage of the task.} Table~\ref{tab:case-ours} lists the second run step by step. At the first step, with an empty record, the navigator supplies the generic \texttt{heat-object-with-appliance} skill. As soon as the record shows the apple on the countertop, the selection switches to \texttt{object-locator}, and the executor picks up the apple and walks to the microwave. Once the record shows that the executor holds the apple and stands at a closed microwave, the selection switches to \texttt{object-heater} and \texttt{appliance-navigator}, whose workflow is exactly ``open the appliance if closed, then \texttt{heat \{object\} with \{appliance\}}''. The executor opens the microwave and heats the apple on the first attempt. After the record contains the successful heating and the opened fridge, the selection moves on to \texttt{open-receptacle}, and the executor places the apple. The skill set therefore changes at the three points where the task changes stage (the apple is found, the closed microwave is reached, and the fridge is opened), and stays fixed in between.

\begin{table*}[!t]
\centering
\caption{\textbf{Step-by-step trace of \textsc{JitSkill} on an unseen ALFWorld game} (``put a hot apple in fridge'', run 2). Injected skills are those delivered before the executor's action at each step (prefix \texttt{alfworld-} omitted); bold marks a change of selection.}
\label{tab:case-ours}
\footnotesize
\setlength{\tabcolsep}{3pt}
\begin{tabularx}{\textwidth}{@{}c>{\raggedright\arraybackslash}p{0.34\textwidth}>{\raggedright\arraybackslash}p{0.24\textwidth}>{\raggedright\arraybackslash}X@{}}
\toprule
Step & Injected skills & Executor action & Observation \\
\midrule
1 & heat-object-with-appliance & go to countertop 1 & sees apple 1 and apple 2 \\
2 & \textbf{object-locator} & take apple 1 from countertop 1 & picks up apple 1 \\
3 & object-locator & go to microwave 1 & the microwave is closed \\
4 & \textbf{object-heater, appliance-navigator} & open microwave 1 & the microwave is open \\
5 & appliance-navigator, object-heater & heat apple 1 with microwave 1 & ``You heat the apple 1'' \\
6 & object-heater, appliance-navigator & go to fridge 1 & the fridge is closed \\
7 & appliance-navigator, object-heater & open fridge 1 & the fridge is open \\
8 & \textbf{heat-object-with-appliance, open-receptacle} & move apple 1 to fridge 1 & task completed \\
\bottomrule
\end{tabularx}
\end{table*}

\paragraph{The same stage transitions recur across runs.} In runs 1 and 3 the executor first visits the fridge before looking for the apple. In both runs, the navigator then supplies \texttt{object-locator} while the apple is being picked up and \texttt{temperature-regulator} once it is in hand. When the executor reaches the closed microwave, the selection again centers on \texttt{object-heater}, paired with \texttt{appliance-navigator} or \texttt{appliance-preparer}, until the fridge is opened, after which both runs switch to \texttt{heat-object-with-appliance} and \texttt{open-receptacle} for placement. Although the action order differs between runs, the skills track the recorded progress rather than the static task description.

\paragraph{Without progress-aware support the executor loops at the heating stage.} The no-skill executor reaches the same intermediate state in all three runs: it finds the apple and walks to the microwave. It then moves the apple into the microwave, closes the door, and issues \texttt{use microwave 1}, instead of heating the apple while holding it. Because the apple is no longer in hand, every later \texttt{heat} command returns ``Nothing happens'', and the executor cycles through the stove burners, the toaster, and the coffee machine until the step limit. In run 1 it finally places the unheated apple in the fridge, which fails the task. The \texttt{object-heater} skill that \textsc{JitSkill} injects at exactly this stage states the correct action form and names the failure directly: ``Nothing happened'' on heat: check (1) you are holding the object.

Figure~\ref{fig:case-block} shows the complete block injected at step 5 of run 2, consisting of the explicit progress record followed by the two selected skills.

\endgroup

\begin{plainpromptbox}{Injected block at step 5 (run 2): progress record and the two selected skills}
Progress notes for the current episode (observed, not instructions):
GOAL: put a hot apple in fridge.
ACTIONS DONE (4 so far, latest last):
- go to countertop 1 -> You arrive at countertop 1. On the countertop 1, you see a apple 2, a apple 1, a mug 3, a ...
- take apple 1 from countertop 1 -> You pick up the apple 1 from the countertop 1.
- go to microwave 1 -> You arrive at microwave 1. The microwave 1 is closed.
- open microwave 1 -> You open the microwave 1. The microwave 1 is open. In it, you see a mug 1.

REFERENCE ONLY -- skill documents from a public skill library, written for other episodes. Nothing below has been executed in the current episode and any values in it are examples, not observations. Use it as guidance for what to do next; you must still act through your own actions.

[SKILL] alfworld-appliance-navigator
Navigates the agent to a target appliance (microwave, stove, fridge, or sinkbasin) needed for object processing. Use when you are holding an object that needs heating, cooling, or cleaning and must move to the correct appliance station. Identifies the required appliance from the task context and executes the movement action.
# Instructions
Use this skill when you are holding an object that needs to be processed (heated, cooled, or cleaned) and you must locate and move to the correct appliance to perform the action.
## Workflow
1. **Identify the Target Appliance:** Determine which appliance is required for the task. Map the action to the appliance: `heat` -> microwave/stoveburner, `cool` -> fridge, `clean` -> sinkbasin.
2. **Locate the Appliance:** Scan the environment observation for the target appliance (e.g., `microwave 1`, `fridge 1`, `sinkbasin 1`).
3. **Navigate:** Execute `go to {appliance}` to move to the identified appliance location.
4. **Prepare Appliance (if needed):** Upon arrival, check if the appliance requires preparation (e.g., opening a closed microwave or fridge door). If so, perform `open {appliance}` before proceeding.
## Example
**Scenario:** You are holding `potato 1` and need to heat it.
  Thought: I need to heat this potato. The microwave is the appropriate appliance.
  Action: go to microwave 1
  Observation: The microwave 1 is closed.
  Thought: I need to open the microwave before I can use it.
  Action: open microwave 1
  Observation: You open the microwave 1. The microwave 1 is open. In it, you see nothing.
**Result:** You are now at the open microwave, ready to heat the potato.
## Key Principles
- **Trigger:** The agent is holding an object and the next step in the task is to `heat`, `cool`, or `clean` it.
- **Core Action:** The primary output of this skill is the navigation command `go to {target_appliance}`.
- **Prerequisite Check:** Always ensure the appliance is accessible (e.g., open) before attempting to use it for processing.

[SKILL] alfworld-object-heater
Heats a specified object using an available heating appliance (e.g., microwave, stoveburner). Use when you are holding an object that requires heating and need to navigate to and operate the heating appliance. Takes the object and appliance as inputs and results in the object being in a heated state.
# Instructions
Heat an object you are holding using a compatible heating appliance (microwave or stoveburner).
## Prerequisites
- The target object must be in your inventory
- A heating appliance (microwave, stoveburner) must exist in the environment
## Workflow
1. **Navigate:** `go to {appliance}` (e.g., `go to microwave 1`)
2. **Check state:** If observation says appliance is closed, execute `open {appliance}` -- verify observation confirms it is now open
3. **Heat:** `heat {object} with {appliance}` -- verify observation confirms "You heat the {object}"
4. **(Optional) Place:** If the task requires it, navigate to the destination and place the heated object
## Action Format
- `go to microwave 1` / `go to stoveburner 1`
- `open microwave 1`
- `heat {object} with {appliance}` (e.g., `heat potato 1 with microwave 1`)
## Error Recovery
- "Nothing happened" on heat: Check (1) you are holding the object, (2) appliance is open, (3) appliance name is correct
- Appliance occupied: Take the existing item out first, then retry
## Example
**Scenario:** You are holding `potato 1` and need to heat it.
  Thought: I need to heat the potato. I'll go to the microwave.
  Action: go to microwave 1
  Observation: The microwave 1 is closed.
  Action: open microwave 1
  Observation: You open the microwave 1. The microwave 1 is open. In it, you see nothing.
  Action: heat potato 1 with microwave 1
  Observation: You heat the potato 1 using the microwave 1.
**Result:** The potato is now heated. Proceed to place it at the destination if required by the task.
\end{plainpromptbox}
\noindent\begin{minipage}{\columnwidth}
\captionof{figure}{\textbf{Complete skill block injected at step 5 of the case-study episode} (run 2). The explicit progress record precedes the reference header and the two selected skills; the whole block fits within the per-step budget and is appended to the latest observation.}
\label{fig:case-block}
\end{minipage}

\FloatBarrier
\section{Prompts}
\label{app:prompts}

This section lists the prompts used in the experiments. All methods share the executor prompt of each environment, and retrieved skills are delivered only through the injected block, so the methods differ only in what that block contains.

\paragraph{Executor prompt.} On ALFWorld, the executor receives the system prompt below, followed by the initial observation, which contains the task goal, as the first user message. Each later observation is returned as a user message of the form \texttt{Observation: \ldots}, and the action is parsed from the \texttt{Action:} line of the response. No few-shot demonstrations are used.

\begin{plainpromptbox}{ALFWorld executor system prompt}
Interact with a household to solve a task. Imagine you are an intelligent agent in a household environment and your target is to perform actions to complete the task goal. At the beginning of your interactions, you will be given the detailed description of the current environment and your goal to accomplish.
For each turn, think briefly and then choose exactly one of these two outputs:
1. `Thought: ...` followed by `Action: ...` for a normal environment action.
2. `Thought: ...` followed by `SkillRequest: ...` if you want extra skill help before acting.

Use `SkillRequest` when it directly helps the next step, especially if you are blocked, the action syntax is unclear, the retrieved skill names look unrelated to the current blocker, or 1-2 actions already failed. In GoS mode, prefer `SkillRequest: GOS_RETRIEVE ...` when you need a better-targeted skill search. In vector mode, prefer `SkillRequest: VECTOR_RETRIEVE ...` when you need a better-targeted vector-only search. Use `READ_SKILL` only when you already know which exact skill is worth reading. Do not output both `Action:` and `SkillRequest:` in the same turn.

The available actions are:
1. go to {recep}
2. take {obj} from {recep}
3. move {obj} to {recep}
4. open {recep}
5. close {recep}
6. use {obj}
7. clean {obj} with {recep}
8. heat {obj} with {recep}
9. cool {obj} with {recep}
where {obj} and {recep} correspond to objects and receptacles.
After your each turn, the environment will give you immediate feedback based on which you plan your next few steps. if the envrionment output "Nothing happened", that means the previous action is invalid and you should try more options.

Your response should use one of the following formats:

Thought: <your thoughts>
Action: <your next action>

or

Thought: <your thoughts>
SkillRequest: <your tool-style request>
\end{plainpromptbox}

\paragraph{Injected skill block of \textsc{JitSkill}.} At every step, the block below is appended to the latest observation on ALFWorld, or to the latest tool result on SkillsBench, immediately before the executor's turn. On $\tau^2$-Bench, the same block is appended to the end of the system prompt, because executors tend to read a block placed in the last user turn as words of the simulated customer. The progress record is produced by deterministic rules without any model call: the goal is taken from the task statement (at most $400$ characters), each executed action is stored with its observation truncated to $90$ characters, and the most recent six actions and six failed actions are rendered. Skill documents are packed in the navigator's ranking order. On SkillsBench, each skill is injected as a compact pointer (name, one-line purpose, and mounted path to the full \texttt{SKILL.md}).

\begin{plainpromptbox}{Template of the injected block}
Progress notes for the current episode (observed, not instructions):
GOAL: <task goal>
ACTIONS DONE (<n> so far, latest last):
- <action> -> <observation>
...
FAILED ACTIONS:
- <action> -> <error observation>

REFERENCE ONLY -- skill documents from a public skill library, written for other episodes. Nothing below has been executed in the current episode and any values in it are examples, not observations. Use it as guidance for what to do next; you must still act through your own actions.

<skill document 1>

<skill document 2>

<skill document 3>
\end{plainpromptbox}

\paragraph{Baseline prompts.} Baselines that maintain or summarize state call the same frozen executor with greedy decoding and thinking disabled. The prompts below are those used on ALFWorld; the prompts on the other benchmarks differ only in environment-specific vocabulary.

\begin{plainpromptbox}{SGDR: state summarizer (the two examples are descriptions drawn from the skill library)}
You are a state summarizer for an agent whose skill library is indexed by descriptions like '<example 1>' or '<example 2>'. Your summary will be cosine-matched against such skill descriptions, so use the SAME operational vocabulary they do.

Given the agent's latest action and the environment observation it produced, produce ONE short paragraph (1-2 sentences) that:
1. Names the kind of state the agent is in, in operational terms (what it has just obtained, where it is, or what error it hit).
2. Lists the action verbs this state ENABLES right now -- i.e. what sub-routines could plausibly run on this exact state. Use verb + object phrasing.

Do NOT enumerate every visible element or value, do NOT describe pure formatting, and do NOT mention task instructions or speculate about future steps. Output only the summary text.
\end{plainpromptbox}

\begin{plainpromptbox}{FlowState: active-state update}
You maintain the Active State of an agent that is executing a task (FlowState). The Active State is a set of typed nodes. Node types: Preference (preferences and constraints the task imposes), Knowledge (facts observed from the environment or tool outputs), Attribute (attributes of the entities involved), Judgement (the agent's conclusions and decisions so far), Artifact (things the agent has produced or changed). Every node has an id, a type, a one-line summary and a value.
Given the task, the current Active State and the latest action and observation, output ONLY a JSON object with the minimal state change for the current request:
{"state_delta": {"add": [{"id": "K1", "type": "Knowledge", "summary": "...", "value": "..."}], "update": [{"id": "...", "summary": "...", "value": "..."}], "remove": ["id"]}}
Use short ids (first letter of the type + a number). Do not repeat unchanged nodes. Remove nodes that are no longer true.
\end{plainpromptbox}

\begin{plainpromptbox}{StructAgent: requirement decomposition and verifier}
[Decomposition] Decompose the task into the ordered list of required outcomes that must ALL hold for the task to be complete. Each requirement is one concrete, verifiable outcome in the environment. Do not add requirements the task does not imply. Output ONLY JSON: {"requirements": ["...", "..."]}

[Verifier] You are the verifier of an agent's progress. Given the task, the requirement table (id, status, requirement) and the agent's latest action and observation, decide which requirements the observation now VERIFIES (direct evidence the outcome holds) and which it INVALIDATES (evidence the outcome no longer holds or was undone). Change a status only on direct evidence in the observation. Output ONLY JSON listing the changed requirements: {"updates": [{"id": "R1", "status": "Verified", "evidence": "..."}]} with status one of Verified, Invalidated, Pending. Output {"updates": []} if nothing changed.
\end{plainpromptbox}

\begin{plainpromptbox}{CodeSkill: skill manager}
You are the skill manager of a household-robot agent (ALFWorld). You maintain a library of reusable skills learned from experience. Each skill has: title; granularity ("task" = a strategy for a whole task family, "event" = how to react to a specific local situation such as an error message, a closed receptacle, or an object not found); when_to_apply (a precise trigger condition); rules (short, concrete, imperative rules using the environment's exact action syntax, e.g. 'go to {recep}', 'take {obj} from {recep}', 'move {obj} to {recep}', 'heat {obj} with {recep}'). Skills must generalise: never mention object numbers or room-specific details.
Given one finished episode and the related existing skills, output maintenance operations as JSON:
{"operations": [ ... ]} where each operation is one of
  {"op": "ADD", "skill": {"title": ..., "granularity": "task"|"event", "when_to_apply": ..., "rules": [...]}}
  {"op": "MERGE", "target": "<existing id>", "skill": {<the full updated skill>}}
  {"op": "DROP", "target": "<existing id>"}
ADD only knowledge that is new and useful; MERGE when an existing skill covers the same situation and the episode adds or corrects rules; DROP an existing skill only if the episode shows it is wrong or misleading. Use at most 3 operations; an empty list is allowed. Output only the JSON object.
\end{plainpromptbox}

\end{document}